\documentclass{article}

\PassOptionsToPackage{numbers,super}{natbib}

\usepackage[preprint]{neurips_2026}

\usepackage[utf8]{inputenc} 
\usepackage[T1]{fontenc}    
\usepackage{hyperref}       
\usepackage{url}            
\usepackage{booktabs}       
\usepackage{amsfonts}       
\usepackage{nicefrac}       
\usepackage{microtype}      
\usepackage{xcolor}         

\usepackage{amsmath}
\usepackage{graphicx}
\usepackage{listings}
\usepackage{multirow}
\usepackage[section]{placeins}
\title{Graph Machine:\\Exploring Edge Mechanisms as an Inductive Bias}

\author{
  \href{https://lintaihou.com}{Lintai Hou}\\
  \texttt{lintai@iterlabs.ai}\\
}

\begin{document}

\maketitle

\begin{abstract}
  Transformers provide a powerful architecture for global content-based matching, but reasoning problems may benefit from a stronger inductive bias toward iterative traversal of latent relations.
  We introduce Graph Machine, an architecture with two explicit edge-based mechanisms:
  Edge-augmented attention, in which edges modulate attention between nodes,
  and edge-centric referral, in which nodes exchange addresses to update their edges.
  Conceptually, this enables the model to dynamically and differentiably construct and revise relational graphs across layers.
  We study this inductive bias using Sudoku under controlled settings and find that Graph Machine outperforms Transformer baselines,
  with ablation studies and mechanistic analysis attributing the gains to the edge mechanisms.
  Surprisingly, we found that the model discovers a compact edge-based construction for Sudoku geometry.
  Our results support explicit edge mechanisms as a promising architectural design, motivating broader evaluation.
\end{abstract}

\begin{figure}[!htbp]
  \centering
  \includegraphics[width=0.96\linewidth]{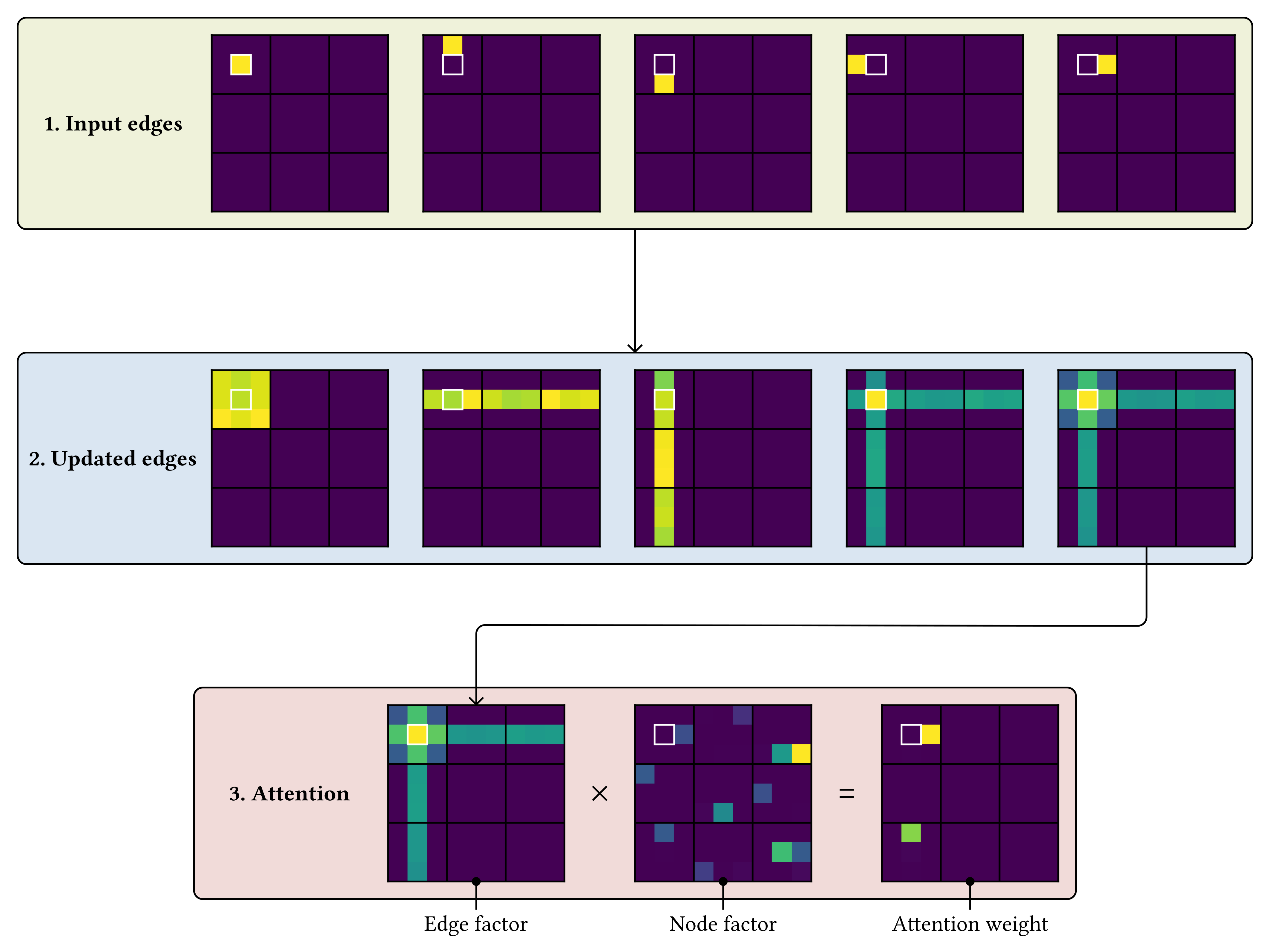}
  \label{fig:overview}
\end{figure}

\section{Introduction}

Most ML architectures, including Transformers~\citep{transformer} and graph neural networks (GNNs)~\citep{gnn,gat}, can be viewed as operating over nodes that aggregate information from other nodes before applying node-wise operations.
This view raises a question: from which other nodes should a node absorb information?
Modern architectures commonly answer this question through some combination of global content-based matching and iterative relational traversal.
The former is exemplified by dot-product self-attention, which selects nodes based on their features.
The latter, by contrast, relies on edges, whether explicit or implicit, to propagate information through a relational structure.
A directed edge can be understood as carrying the addresses of its target nodes, and potentially also features that describe the relation between the source and the target.

We hypothesize that the relative ease of global content-based matching versus iterative relational traversal constitutes an important inductive bias in architectural design.
Systems like large language models (LLMs)~\citep{llm} are known to favor surface heuristics over latent constraints~\citep{car-wash-paper}.
Among many examples of what is described as shortcut learning, one demonstrates it with particular clarity: ``I want to wash my car. The car wash is 50 meters away. Should I walk or drive?''
An LLM often answers ``walk'' by treating the problem as a familiar distance-focused question and attending to the salient distance cue.
The correct answer, however, is to drive, because the car itself must be brought to the car wash~\citep{car-wash-post}.
Global content-based matching gives a model broad reach and rich representational capacity, but it also makes shallow shortcuts immediately available.
Iterative relational traversal, by contrast, may be better aligned with reasoning over latent relations and may regularize against content-based shortcuts.
If this prior holds, then making the latter easy becomes a central desideratum.

However, we argue that architectures like Transformers and GNNs often do the opposite.
One important reason is the lack of an edge-centric referral mechanism.
Such a mechanism would encourage the model to pass not only messages between nodes, but also addresses.
These addresses could then be used to construct new edges for the next information aggregation step.
Intuitively, this allows nodes to fetch a neighbor's neighbor by letting the neighbor provide the referral.

GNNs commonly model edges directly by providing each node with a set of neighbors, which the node then uses for message passing.
However, the topology is typically fixed, or dynamic only through a procedure that is external to the model's own computation.
A GNN may be forced to operate over a graph structure that is not amenable to the computation the model wishes to perform,
and therefore suffers from linearly depth-limited receptive fields, 1-WL expressivity~\citep{1-wl} bound, and oversquashing~\citep{oversquashing}.

By contrast, dot-product self-attention in Transformers directly supports global content-based matching.
Although self-attention may have enough capacity to implement iterative relational traversal and referral, it makes these operations difficult in three ways:
limited addressing precision, noisy address channels, and competition with global content-based matching.

First, dot-product attention retrieves nodes through dot products in a $d_k$-dimensional key/query space.
Viewing exact node addresses as one-hot vectors in an $n$-dimensional address space, this can be interpreted as preserving address distinguishability in a $d_k$-dimensional geometry,
suggesting a trade-off between dimension and distortion analogous to Johnson--Lindenstrauss-type bounds~\citep{jl}.

Second, fetched addresses must be carried through the value stream and subsequent feed-forward networks.
These transformations are useful for representation learning, but noisy for exact bookkeeping: an address should ideally be forwarded without modification.

Third, global content-based matching and iterative relational traversal/referral share the same infrastructure, forcing them to compete for dimensions, heads, and optimization pressure.
Thus, the balance between the two modes of computation is determined by implicit trade-offs, leaving no explicit control to bias the model toward the latter.

This work introduces Graph Machine (GM), an architecture that directly supports edge-centric referral while generalizing Transformers in expressivity.
Graph Machines update node features through edge-augmented attention and construct new edges through edge-centric referral.

In controlled Sudoku experiments, Graph Machines outperform same-scale and prior-advantaged Transformer baselines, while remaining comparable to substantially enlarged and prior-advantaged Transformer baselines.
Mechanistic analysis suggests that this advantage reflects the intended inductive bias.
Starting from only local adjacency edges, GM learns to construct higher-level Sudoku relations, including box, row, and column regions, through a compact edge-based construction that resembles a $1$-$2$-$4$ expansion process.
This provides evidence that the model is not merely fitting Sudoku solutions, but using referral to build task-relevant latent graph structure.

These results suggest that explicit edge mechanisms can provide a useful inductive bias for structured reasoning tasks.

\section{Architecture}

\begin{figure}[!htbp]
  \centering
  \begin{minipage}{0.48\textwidth}
    \centering
    \includegraphics[width=\linewidth]{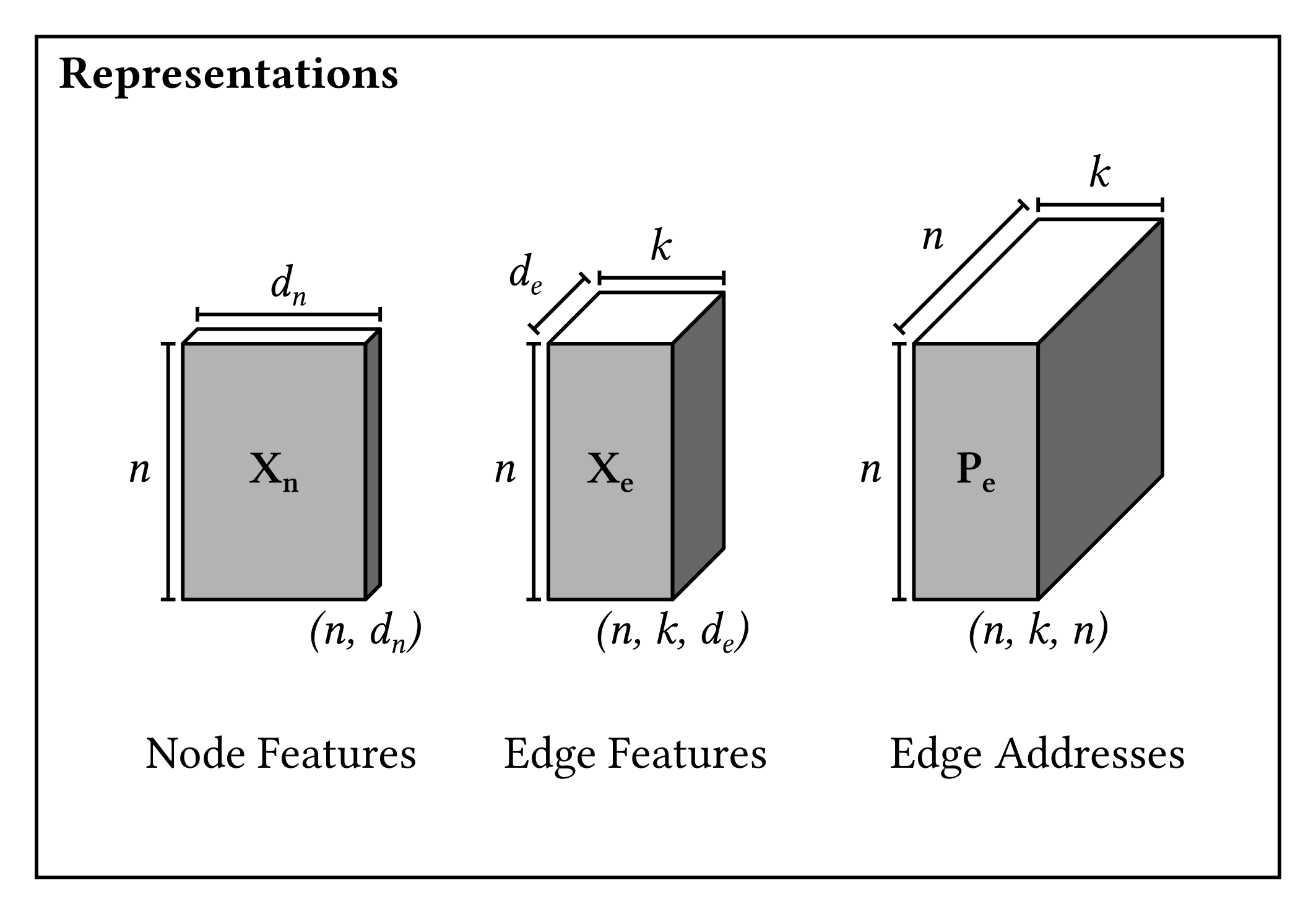}
    \caption{GM representations.}
    \label{fig:representations}
  \end{minipage}
  \hfill
  \begin{minipage}{0.48\textwidth}
    \centering
    \includegraphics[width=\linewidth]{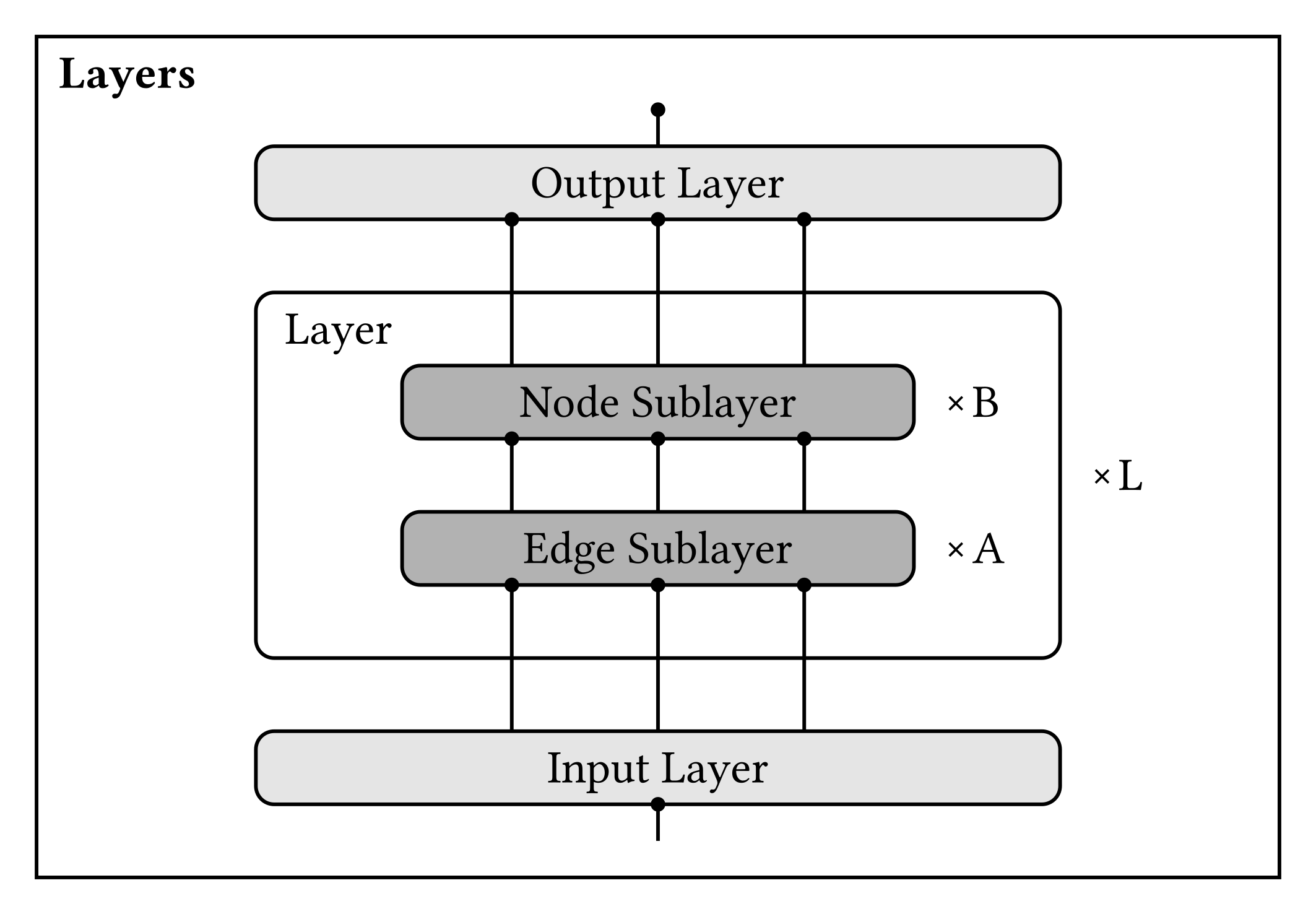}
    \caption{GM layers.}
    \label{fig:layers}
  \end{minipage}
\end{figure}

\begin{figure}[!htbp]
  \centering
  \begin{minipage}{0.48\textwidth}
    \centering
    \includegraphics[width=\linewidth]{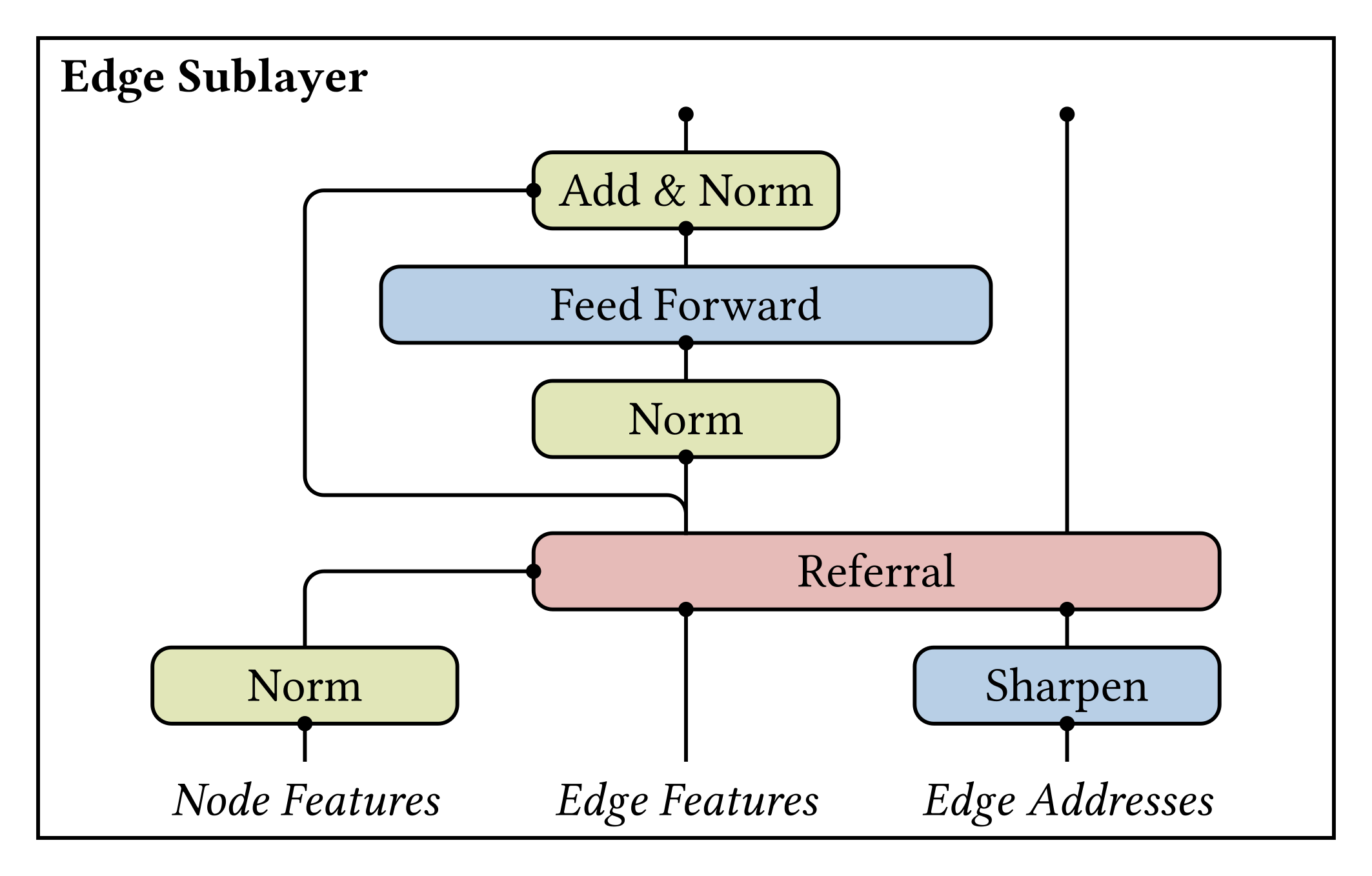}
    \caption{GM edge sublayer.}
    \label{fig:edge-sublayer}
  \end{minipage}
  \hfill
  \begin{minipage}{0.48\textwidth}
    \centering
    \includegraphics[width=\linewidth]{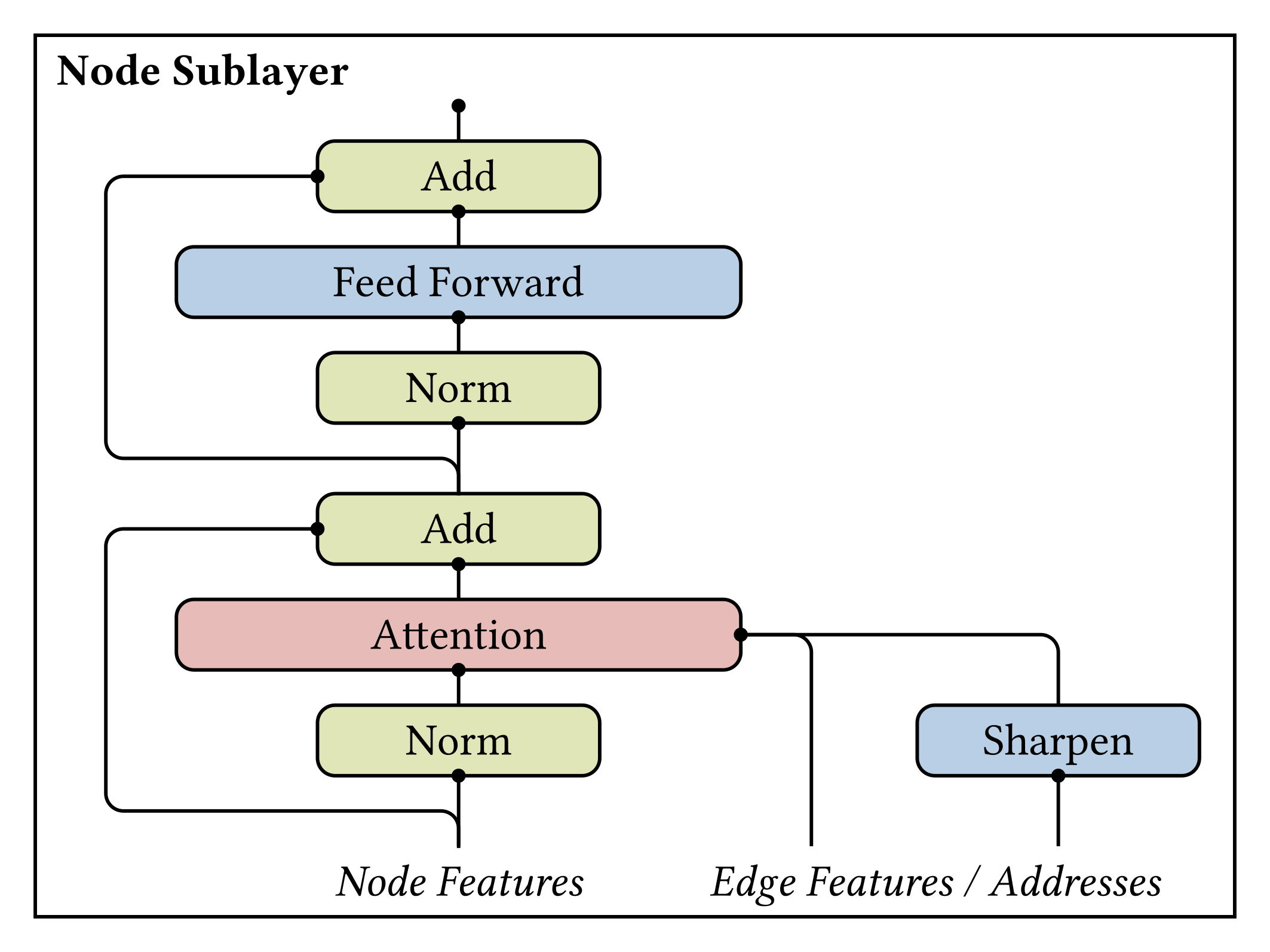}
    \caption{GM node sublayer.}
    \label{fig:node-sublayer}
  \end{minipage}
\end{figure}

\subsection{Architecture overview}

We conceptualize a Graph Machine as operating on a graph with $n$ nodes, where each node maintains $k$ outgoing edge slots (Figure~\ref{fig:representations}).
In addition to node features of shape $n\times d_n$, the model maintains edge features of shape $n\times k\times d_e$ and edge addresses of shape $n\times k\times n$, where $d_n$ and $d_e$ denote the node and edge hidden sizes, respectively.
The edge addresses represent target mass: each edge assigns weights over the $n$ target nodes.

The initial edge features and edge addresses are provided by the input layer, either by converting task-specific input graphs or through priors such as $k$ past neighbors in the case of sequence modeling (Figure~\ref{fig:layers}).
Each intermediate layer of a Graph Machine consists of one or more edge sublayers followed by one or more node sublayers, allowing many-to-one or one-to-many arrangements.
Edge sublayers update edge representations using edge-centric referral, followed by a position-wise feed-forward network that updates edge features (Figure~\ref{fig:edge-sublayer}).
Node sublayers update node features using edge-augmented attention followed by their own position-wise feed-forward network (Figure~\ref{fig:node-sublayer}).

Viewed through a programming analogy, node features act like an object's general state, while each edge slot resembles a relational field that stores both relation-specific state and a soft pointer.
Each object can traverse its pointers to reach target objects, fetch their states and pointers, and use the retrieved information to revise its own pointers while updating its general and relational states.

Another way to view Graph Machines is through relational composition.
For an ordinary graph with adjacency matrix $P$, the composition operator $\circ$ is defined by matrix multiplication: $P\circ P = P^2$.
Graph Machines generalize relational representation from connectivity to multiple feature-bearing soft edges.
The edge representation is a pair $(X_e, P_e)$, where $X_n$ and $X_e$ stores node and node-relation-specific features and $P_e$ is edge addresses, equivalently $k$ row-normalized soft adjacency matrices.
Edge-centric referral then implicitly defines a learned differentiable composition operator $\diamond_\theta$ over this representation: $(X'_e, P'_e) = \diamond_\theta(X_n, X_e, P_e)$.

\begin{figure}[!htbp]
  \centering
  \includegraphics[width=0.96\linewidth]{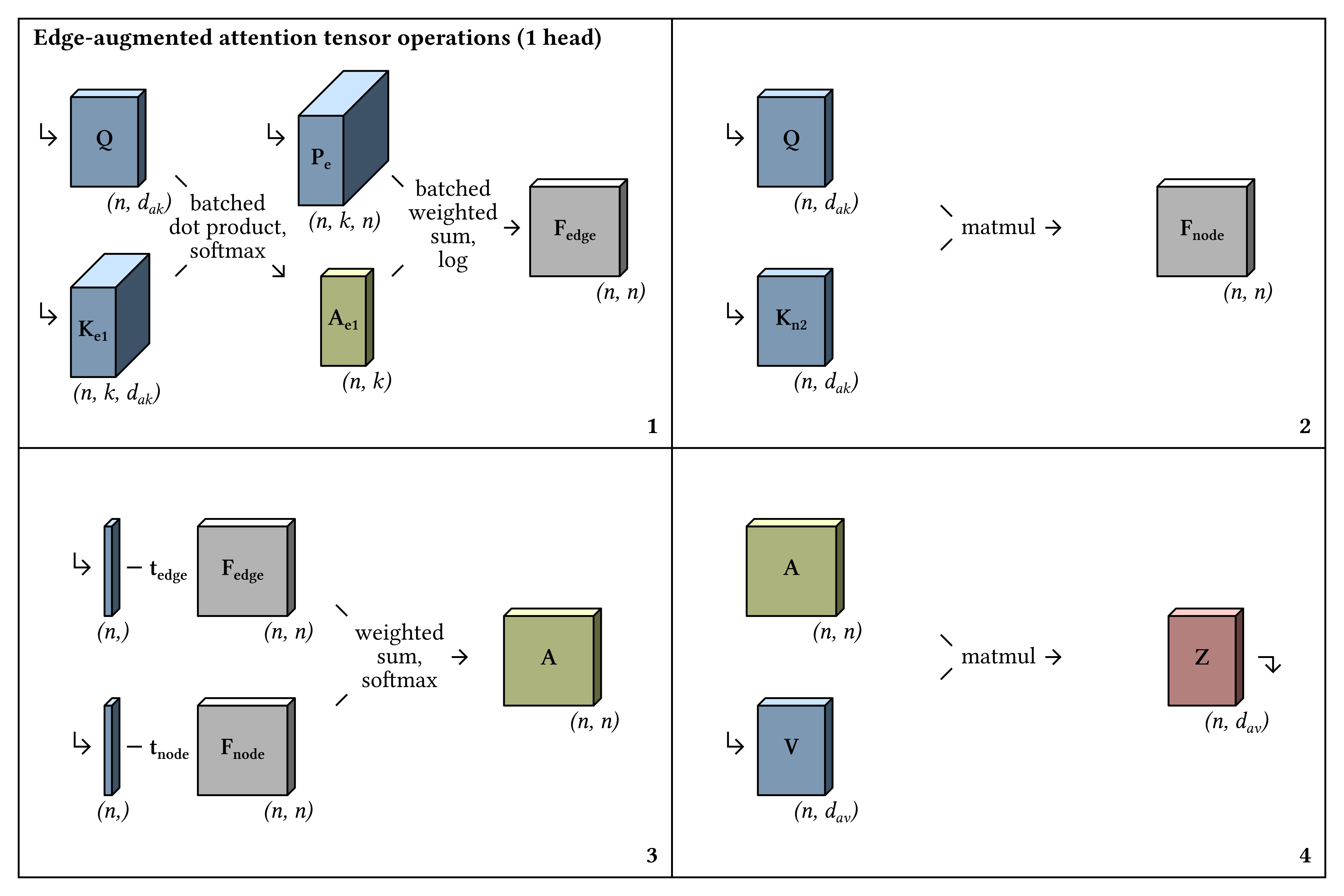}
  \caption{Edge-augmented attention tensor operations  ($1$ head).}
  \label{fig:attention}
\end{figure}

\subsection{Edge-augmented attention}

Standard multi-head dot-product self-attention computes attention logits from queries and keys projected from node features ($X_n$).
Graph Machines retain this term as a \emph{node factor} ($F_{\mathrm{node}}$), but augment it with an \emph{edge factor} ($F_{\mathrm{edge}}$) derived from edge features ($X_e$) and edge addresses ($P_e$).
The final attention weights are a product-of-experts~\citep{poe} combination between the factors, so that node-based content affinity and edge-based relational evidence jointly shapes the target map.

For simplicity, we henceforth omit the head dimension and denote a node-edge-node-edge-node chain by $(n1, e1, n2, e2, n3)$,
where $n1$ is the source node, $e1$ is an edge of the source node, $n2$ is a target node, and $e2$ is an edge of the target node.
Queries ($Q$) are projected from $n1$ features, while the standard attention keys are projected from $n2$ features; under the chain notation above, we refer to them as \emph{$n2$ keys} ($K_{n2}$).
To produce the edge factor, the same $n1$ queries attend over \emph{$e1$ keys} ($K_{e1}$), which are projected from the $k$ edge features associated with $e1$.
The corresponding edge addresses are used as values, yielding a mixture of edge-address distributions over target nodes ($A_{e1}$).
To convert this mixture into logit space, we then take the logarithm of this target mass after clipping away from $0$ with a small $\varepsilon$.
Concretely,
\[
  \begin{aligned}
    A_{e1}(n1,e1)={}            & \operatorname{softmax}_{e1}\!\Big(\frac{Q(n1)\,K_{e1}(n1,e1)^\top}{\sqrt{d_{\mathrm{ak}}}}\Big), \\
    M_{\mathrm{edge}}(n1,n2)={} & \sum\nolimits_{e1}A_{e1}(n1,e1)\,P_e(n1,e1,n2),                                                  \\
    F_{\mathrm{edge}}(n1,n2)={} & \log\!\Big(\!\max\!\big(M_{\mathrm{edge}}(n1,n2),\,\varepsilon\,\big)\Big),                      \\
    F_{\mathrm{node}}(n1,n2)={} & \frac{Q(n1)\,K_{n2}(n2)^\top}{\sqrt{d_{\mathrm{ak}}}}.
  \end{aligned}
\]

The node and edge factors are multiplied by learned temperature scalars projected from node features and parameterized by a positive function such as $\exp$ or $\operatorname{softplus}$ ($t_{\mathrm{node}}$ and $t_{\mathrm{edge}}$).
This allows the model to modulate the relative strength of the two experts.
Summing the $2$ factors and applying a softmax over the target-node dimension gives the final attention weights.
The mechanism is at least as expressive as standard Transformer attention: vanilla self-attention is recovered by setting the edge-factor temperature to $0$, so that the edge expert contributes a uniform distribution.
Concretely,
\[
  \begin{aligned}
    A(n1,n2)={} & \operatorname{softmax}_{n2}\!\Big(t_{\mathrm{node}}(n1)\,F_{\mathrm{node}}(n1,n2)+t_{\mathrm{edge}}(n1)\,F_{\mathrm{edge}}(n1,n2)\Big), \\
    Z(n1)={}    & \sum\nolimits_{n2}A(n1,n2)\,V(n2).
  \end{aligned}
\]

We visualize the tensor operations during one head of the edge-augmented attention mechanism in Figure~\ref{fig:attention} and provide simplified code in Section~\ref{sec:simplified-code-attention}.

\begin{figure}[!htbp]
  \centering
  \includegraphics[width=0.96\linewidth]{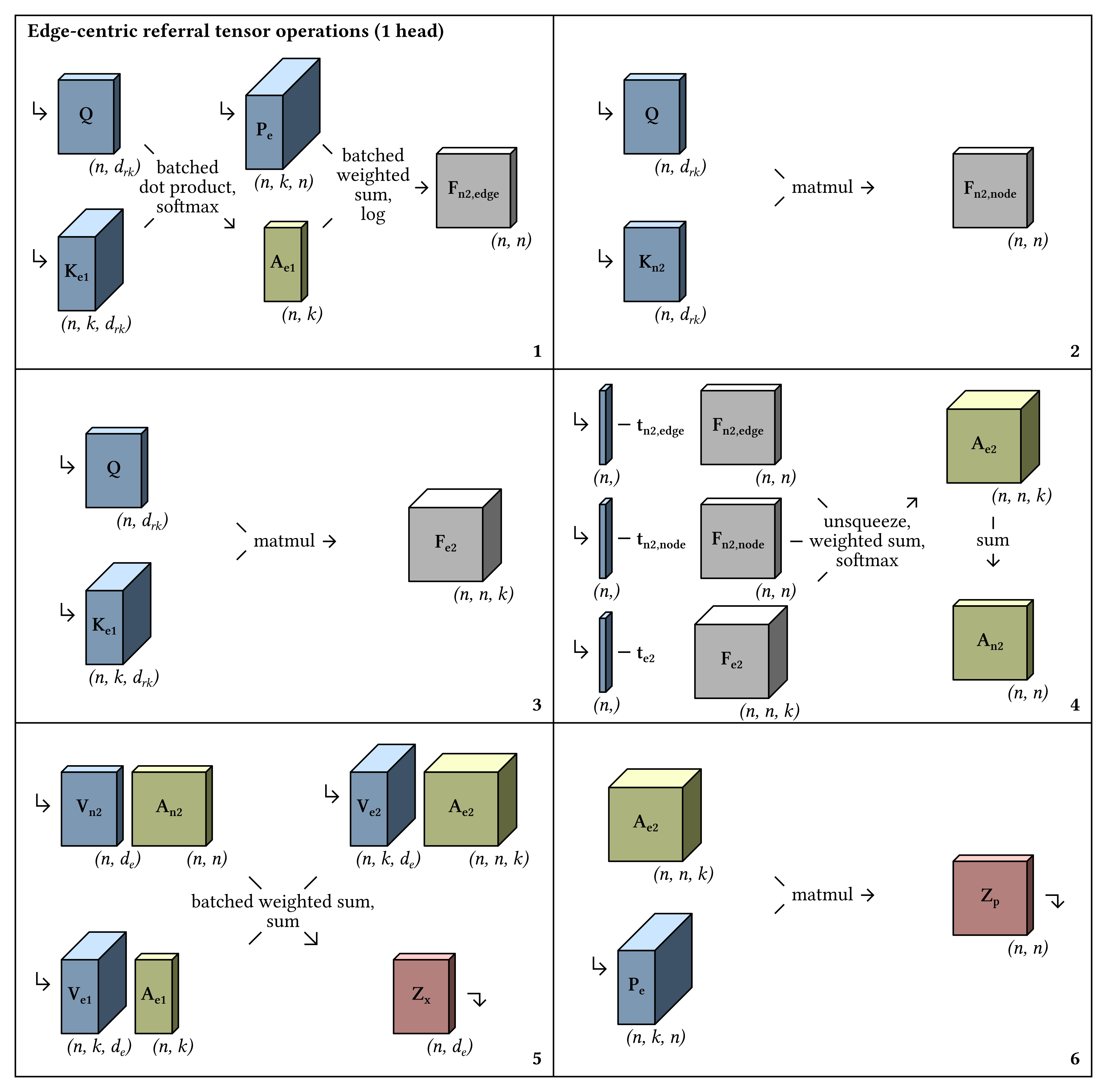}
  \caption{Edge-centric referral tensor operations ($1$ head).}
  \label{fig:referral}
\end{figure}

\subsection{Edge-centric referral}

Edge-centric referral constructs a new set of edges by composing two-hop $n1 \to n2\to n3$ chains into single-hop $n1\to n3$ edges with new features and addresses.

During edge-centric referral, the $k$ new edges play a dimensional role analogous to attention heads.
Because self-connections are representable in Graph Machines, the mechanism does not preclude preserving existing edges.

Referral shares its early computational pattern with edge-augmented attention.
The queries produce node and edge factors over possible intermediate nodes $n2$, denoted the \emph{$n2$ node factor} ($F_{n2,\mathrm{node}}$) and \emph{$n2$ edge factor} ($F_{n2,\mathrm{edge}}$),
while the attention weights over the original $e1$ edge slots are retained as \emph{$e1$ weights} ($A_{e1}$).
The same queries also attend over the outgoing $e2$ edge slots of each $n2$ to produce an \emph{$e2$ factor} ($F_{e2}$).
Combining the $n2$ node factor, $n2$ edge factor, and $e2$ factor with learned temperatures yields logits over referred edges; applying a softmax gives \emph{$e2$ weights} ($A_{e2}$).
These weights define a distribution over candidate $e2$ edges.
The corresponding \emph{$n2$ weights} ($A_{n2}$) are obtained by summing the $e2$ weights over the $e2$ edge dimension.
By contrast, a different variant would first compute $n2$ weights and then multiply them by softmax-normalized $e2$ factors, thereby modeling a conditional distribution over $e2$ given $n2$.
Concretely, after early analogous operations,
\[
  \begin{aligned}
    F_{e2}(n1,n2,e2)={} & \frac{Q(n1)\,K_{e2}(n2,e2)^\top}{\sqrt{d_{\mathrm{rk}}}},                                               \\
    A_{e2}(n1,n2,e2)={} & \operatorname{softmax}_{n2,\,e2}\!\Big(t_{n2,\mathrm{node}}(n1)\,F_{n2,\mathrm{node}}(n1,n2)            \\
                        & \phantom{\operatorname{softmax}_{n2,\,e2}\!\Big(}+t_{n2,\mathrm{edge}}(n1)\,F_{n2,\mathrm{edge}}(n1,n2) \\
                        & \phantom{\operatorname{softmax}_{n2,\,e2}\!\Big(}+t_{e2}(n1)\,F_{e2}(n1,n2,e2)\Big),                    \\
    A_{n2}(n1,n2)={}    & \sum\nolimits_{e2}A_{e2}(n1,n2,e2).
  \end{aligned}
\]

The new edge features are obtained by taking weighted sums of the values associated with $e1$, $n2$, and $e2$ ($V_{e1}$, $V_{n2}$, and $V_{e2}$), using their respective weights, then adding the results and projecting back to the edge hidden dimension.
The new edge addresses are computed by weighting the addresses of the $e2$ edges with the $e2$ weights.
Consequently, edge addresses would remain convex combinations of previous addresses absent the sharpening procedure.
Concretely,
\[
  \begin{aligned}
    Z_x(n1)={}    & \sum\nolimits_{e1}A_{e1}(n1,e1)\,V_{e1}(n1,e1)           \\
                  & +\sum\nolimits_{n2}A_{n2}(n1,n2)\,V_{n2}(n2)             \\
                  & +\sum\nolimits_{n2,\,e2}A_{e2}(n1,n2,e2)\,V_{e2}(n2,e2), \\
    Z_p(n1,n3)={} & \sum\nolimits_{n2,\,e2}A_{e2}(n1,n2,e2)\,P_e(n2,e2,n3).
  \end{aligned}
\]

We visualize the tensor operations during one head of the edge-centric referral mechanism in Figure~\ref{fig:referral} and provide simplified code in Section~\ref{sec:simplified-code-referral}.

\subsection{Sharpening}

Because Shannon entropy $H(\cdot)$ is concave in the edge address distribution $P_e$,
Jensen's inequality gives
\[
  H\Big(\sum\nolimits_{n2,\,e2}A_{e2}(n1,n2,e2)\,P_e(n2,e2,\cdot)\Big)\ge\sum\nolimits_{n2,\,e2}A_{e2}(n1,n2,e2)\,H\big(P_e(n2,e2,\cdot)\big),
\]
where $A_{e2}(n1,n2,e2) \ge 0$ and $\sum\nolimits_{n2,\,e2} A_{e2}(n1,n2,e2)=1$.
Thus, repeatedly forming weighted averages of edge addresses during referral can dilute the target mass of the resulting edges.

To counteract this effect, Graph Machines apply an edge-sharpening operation at the start of edge and node sublayers.
This operation performs learned temperature scaling on edge addresses:
each edge address is mapped to logit space by taking the logarithm, scaled by a learned temperature ($t_{\mathrm{sharpener}}$) projected from edge features and parameterized by a positive temperature function, and renormalized with a softmax.
Concretely,
\[
  \begin{aligned}
    L_e(n1,e1,n2)={}  & \log\!\Big(\!\max\!\big(P_e(n1,e1,n2),\,\varepsilon\,\big)\Big),                     \\
    P'_e(n1,e1,n2)={} & \operatorname{softmax}_{n2}\!\Big(t_{\mathrm{sharpener}}(n1,e1)\,L_e(n1,e1,n2)\Big).
  \end{aligned}
\]

In edge sublayers, one could use separate sharpener temperatures for the two roles played by edge addresses: one version for producing $n2$ edge factors and another as the $e2$ address being collected.
For simplicity, we use a single sharpener temperature for both roles.

\subsection{Computational optimizations}

A direct implementation of the conceptual algorithm incurs an $O(n^2)$ memory footprint and an $O(n^3)$ compute cost in the number of nodes $n$.
These costs arise from materializing an $n\times h\times n$ weight tensor for each batch element in attention and an $n\times k\times n\times k$ weight tensor in referral,
and performing matrix multiplication between the weight tensor and an $n\times k\times n$ address tensor during referral.
Two natural optimization strategies reduce this cost by reducing the footprint of the edge addresses.

The first direction is to compress edge addresses into a lower-dimensional representation and recover them with a linear decoder during attention and referral.
Because a vanilla compression-decompression process does not respect the simplex constraint,
it is more natural to represent the edge address in logit space instead of weight space (we present the results on edge address space without compression in Section~\ref{sec:gm-edge-address-space}).
This requires interpreting edge-address aggregation steps as also a product-of-experts operation rather than a mixture.
In this setting, the edge factors can be computed, together with the node factors, through dot products between queries concatenated with compressed target edge addresses and keys concatenated with the decoder matrix.
This makes the mechanism compatible with FlashAttention-style kernels~\citep{flash-attention}.

The second direction is to sparsify edge addresses to contain only the top-$s$ entries, storing them in a sparse coordinate format and utilizing sparse operations (\verb+scatter+/\verb+gather+) during referral and attention.
This allows memory footprint and computational cost to both be $O(n)$, as long as edge factor temperatures are enforced away from $0$ so that attention/referral weights inherit the sparsity of edge addresses.
Each edge-address aggregation step requires coalescing, and to maintain sparsity, subsequent selection, either via hard top-$s$ or probabilistic Gumbel-top-$s$.
Additionally, $s$ could potentially differ between various operations.
In contrast to vanilla Transformers with dynamic dense interactions and sparse Transformers with fixed routing, a sparse Graph Machine would provide sparse but dynamic attention.
We present preliminary simulated sparsification results in Section~\ref{sec:gm-address-sparsity}.

One limiting case is where edges become hard and referral amounts to assigning weights to at most $k^2$ neighbor candidates and selecting $k$ of them.
With such hardened edges, optimization becomes more difficult: referral is no longer fully differentiable, and would need reinforcement-learning-style estimators such as policy gradients.
In this sense, GM can be viewed as a soft generalization of hard referral, allowing varying degrees of edge softness so that gradients can flow through the referral process.

A further optimization is to avoid requiring each of the $k$ heads to attend separately to each of the $k$ $e2$ edges of an $n2$ node during referral.
Instead, the model can first aggregate edge addresses for each head using a procedure analogous to $e1$ aggregation, reducing the $n\times k\times n\times k$ memory cost to $n\times k\times n$.

\section{Experiments}

\subsection{Experimental setup}

Our experimental design follows the principle of controlled comparison: we isolate the causal effects of the variables of interest by keeping all other factors fixed.
All GM and Transformer conditions are therefore considered instances of a generalized model class and share the same model and training implementation, with configurations kept at their default except for those relevant to the experiments.
We use a deliberately simple setup and avoid many straightforward optimizations; for example, increasing model size or training steps would improve performance but offer little additional insight.

We use the Sudoku-3M dataset from Kaggle~\citep{sudoku-3m,sudoku-src} as our benchmark, which contains $3$M Sudoku puzzles with $23$ to $26$ clues and varying levels of difficulty.
Controlling the task prior given to the model is crucial, since the prior can strongly influence both the task the model faces and its performance.
At one extreme, Sudoku can be solved entirely by human-engineered symbolic algorithms, without using a neural network.
As we intend to use Sudoku to test a model's capability to use input relations to construct relations amenable to reasoning, relations such as the Sudoku constraint regions should not be made readily available.
Therefore, our standard task prior is defined as local information: for each cell, knowledge of the cell itself and of its four adjacent neighbors.

We use a single model pass for both training and testing.
Each of the $81$ cells is treated as a node ($n=81$), with $8$ edges per node ($k=8$).
In conditions with edges, the initial edges point to the cell itself and its four adjacent neighbors; the remaining edge slots, up to $5$ for corner cells, are assigned to empty edges.
Edge features are initialized from learned embeddings of the edge category: self, up, down, left, right, or empty.
Initial edge addresses place all mass on the target cell for non-empty edges and distribute mass uniformly over all cells for empty edges.
Node features are initialized by adding two learned embeddings: one for the cell content, which is either blank or one of $1,\ldots,9$, and one for the cell position, with each of the $81$ positions treated as a separate category.
Prediction logits are produced from the final node embeddings.
Cross-entropy loss is applied only to logits at blank positions, and then averaged over puzzles and subsequently batches.
We report board accuracy as our main metric, where a board is considered correct if and only if each blank cell is predicted correctly.

We use a tiny model that is relatively narrow and deep, motivated by Sudoku's requirement for multi-step reasoning.
For our default configuration, the node hidden size $d_n$ and edge hidden size $d_e$ are $64$ and $8$, respectively; the number of layers $l$ is $32$, with each layer containing $1$ edge sublayer and $1$ node sublayer.
Both the edge degree, equivalently the number of referral heads, $k$ and the number of attention heads $h$ are $8$, and the referral key size $d_{\mathrm{rk}}$ and the attention key and value sizes $d_{\mathrm{ak}}$ and $d_{\mathrm{av}}$ are all $8$.
The detailed default configuration for all conditions is provided in Section~\ref{sec:default-config}.
In general, we follow current best practices for Transformers.

Each run uses a single NVIDIA GeForce RTX 4090 GPU with $24$ GB of VRAM and generally completes in under $15$ hours.
We use integers starting from $0$ as random seeds, randomizing the model initialization, the train-evaluation-test split, and the epoch shuffling.
Conditions in Section~\ref{sec:main-experiments} use $10$ seeds; all other conditions use $3$ seeds.
Parenthesized conditions in later sections are equivalent to one of the main conditions in Section~\ref{sec:main-experiments} (up to numerical error), so their values are taken from the corresponding $10$-seed averages.

\subsection{Comparison against baselines}
\label{sec:main-experiments}

\begin{table}[!htbp]
  \caption{Comparison against baselines.}
  \label{tab:main-results}
  \centering
  \setlength{\tabcolsep}{3pt}
  \begin{tabular}{lccccc}
    \toprule
    Condition                     & Scale     & Node embed.  & Attention  & \# params & Acc ($\%$, $\mu \pm \sigma\ (\text{max})$) \\
    \midrule
    Transformer                   & $1\times$ & Each         & Node       & $2.12$M   & $71.7\ \pm\ 3.2\ (76.1)$                   \\
    Transformer-static            & $1\times$ & Each         & Node, edge & $2.16$M   & $1.8\ \pm\ 1.9\ (5.1)$                     \\
    Transformer-sin-PE$^\star$    & $1\times$ & Each, sin PE & Node       & $2.12$M   & $81.5\ \pm\ 0.5\ (82.3)$                   \\
    Transformer-sin-PE-2x$^\star$ & $2\times$ & Each, sin PE & Node       & $16.87$M  & $86.3\ \pm\ 2.5\ (88.6)$                   \\
    \midrule
    \textit{GM}                   & $1\times$ & Each         & Node, edge & $2.70$M   & $86.0\ \pm\ 3.7\ (87.9)$                   \\
    \bottomrule
  \end{tabular}
\end{table}

We compare Graph Machines against several Transformer-based baselines.

\paragraph{Transformer}
We include a standard Transformer as one of our baselines.
To maintain symmetry with GM conditions, we retain factor temperature and auxiliary entropy losses even in baseline conditions, where attention contains only the node factor.
Ablations suggest that these choices do not materially affect Transformer performances; see Section~\ref{sec:transformer-sin-pe-2x-factor-temperatures} and~\ref{sec:transformer-sin-pe-2x-entropy-losses}.

\paragraph{Transformer-static}
For the Transformer-static condition, we disable edge referral by removing edge sublayers.
The given input edge embeddings and edge addresses are therefore not updated, but are still used by edge-augmented attention.
This condition remains a generalization of Transformer in expressivity: since empty edges with even mass distribution are supplied,
the model can construct a uniform edge expert, effectively adding a constant term to attention scores, therefore not limiting message passing to the given topology.

\paragraph{Transformer-sin-PE}
For the Transformer-sin-PE condition, we provide the Transformer with $2$D positional information through sinusoidal positional embeddings~\citep{transformer} with base $10{,}000$ at the input layer.
This is the best-performing positional embeddings option for Transformer among sinusoidal, rotary~\citep{rope}, and row-column factorized positional embeddings schemes of various bases (Section~\ref{sec:transformer-pe}).

\paragraph{Transformer-sin-PE-2x}
We include a scaled-up Transformer with $2$D sinusoidal position embeddings as a stronger comparison, using models that are $2\times$ wide in hidden size and head dimension sizes and $2\times$ deep in number of layers.

These conditions provide useful isolation of the relevant factors.
The comparison between GM and Transformer isolates the overall effect of edge representations and mechanisms,
while the comparison between GM and Transformer-static isolates the edge-updating referral mechanism.
Transformer-sin-PE examines the effect of task priors by making $0$ to $8$-hop positional information directly accessible at input, from which the Sudoku constraint regions can be easily derived,
even though this arguably removes much of the challenge of constructing useful relations from given ones, which was our intention when choosing Sudoku as the benchmark and local information as our task priors.
Together with an experiment where we project input edges into the node feature space (Section~\ref{sec:transformer-proj}), they remove or reverse the effect of the $0$ to $1$-hop positional task priors given to GM.
Finally, the scaled-up Transformer-sin-PE-2x provides a generous baseline with advantaged task priors and approximately $8\times$ the parameter count.

Results in Table~\ref{tab:main-results} show that GM outperforms Transformer, Transformer-static, and Transformer-sin-PE, while remaining competitive with Transformer-sin-PE-2x.
The Transformer-static condition yields an average per-cell accuracy of $77.3\%$, moderately lower than Transformer's $96.3\%$, with this discrepancy amplified under the per-board accuracy metric.

Transformer-static's poor performance, together with the first three ablation studies, presents an interesting picture.
Several conditions are equally or more expressive than the models they underperform: Transformer-static relative to Transformer, GM with RoPE relative to GM, and 5-edge-degree GM and 4-edge-sublayer GM relative to Transformer.
The natural interpretation is that our architectural changes introduce inductive biases that can materially move models into different regimes, and that these regimes yield gains only when they are both sufficiently supported and advantageous.
The similar accuracy of GM and the strongest Transformer baseline may therefore reflect two distinct regimes being pushed near the limit under our task and setup, rather than a lack of meaningful effect from inductive bias.

\subsection{Ablation studies}
\label{sec:ablation-studies}

\paragraph{Positional encoding}
We add the best performing sinusoidal and rotary positional embeddings schemes (evaluated on Transformer, Section~\ref{sec:transformer-pe}) to GM.
Results (Table~\ref{tab:positional-encoding}) show that sinusoidal PE appears to slightly improve GM performance to be on par with Transformer-sin-PE-2x.
On the other hand, consistent with our unreported preliminary experiments, RoPE appears to stably decrease GM's performance to Transformer-RoPE level shown in Section~\ref{sec:transformer-pe}.
We theorize that RoPE may compete to supply positional information during attention, resulting in a regime that underperforms relative to when edges are used in that role by the model.

\begin{table}[!htbp]
  \centering
  \begin{minipage}{0.48\textwidth}
    \caption{Positional encoding.}
    \label{tab:positional-encoding}
    \centering
    \begin{tabular}{lc}
      \toprule
      Positional encoding  & Accuracy ($\%$) \\
      \midrule
      Sin PE, base $10000$ & $86.4$          \\
      RoPE, base $10$      & $76.5$          \\
      None (GM)            & $86.0$          \\
      \bottomrule
    \end{tabular}
  \end{minipage}
  \hfill
  \begin{minipage}{0.48\textwidth}
    \caption{Edge degree.}
    \label{tab:edge-degree}
    \centering
    \begin{tabular}{lc}
      \toprule
      Degree            & Accuracy ($\%$) \\
      \midrule
      $0$ (Transformer) & $71.7$          \\
      $5$               & $63.2$          \\
      $8$ (GM)          & $86.0$          \\
      \bottomrule
    \end{tabular}
  \end{minipage}
\end{table}

\paragraph{Edge degree}
The edge degree $k$ controls the sparsity of the edge representation and affects referral compute at rate $k^2$.
In addition to an edge degree of $0$ conceptually represented by the Transformer baseline, we test an edge degree of $5$, the smallest value that can accommodate the five provided non-empty edges (up, down, left, right, self) in the input graph.
Interestingly, lower-degree GM appears to underperform relative to both endpoints (Table~\ref{tab:edge-degree}).

\paragraph{Number of edge sublayers}
The ratio of edge to node sublayers controls the relative amount of computation allocated to message passing and address passing.
The Transformer-static condition can be viewed as the limiting case with no edge sublayers, so edge representations are never updated.
We test the effect of varying the number of edge sublayers while fixing the number of node sublayers to $32$.
Performance decreases with fewer edge sublayers, with a trend of diminishing returns toward the upper range (Table~\ref{tab:num-edge-sublayers}).
The average for the $16$-edge-sublayer condition is skewed by seed $1$'s outlier performance of $51.5\%$, compared with $79.1\%$ and $85.9\%$ for seeds $0$ and $2$, respectively.

\begin{table}[!htbp]
  \centering
  \begin{minipage}{0.48\textwidth}
    \caption{Number of edge sublayers.}
    \label{tab:num-edge-sublayers}
    \centering
    \begin{tabular}{lc}
      \toprule
      \# edge sublayers        & Accuracy ($\%$) \\
      \midrule
      $0$ (Transformer-static) & $1.8$           \\
      $4$                      & $59.7$          \\
      $8$                      & $83.3$          \\
      $16$                     & $72.2$          \\
      $32$ (GM)                & $86.0$          \\
      \bottomrule
    \end{tabular}
  \end{minipage}
  \hfill
  \begin{minipage}{0.48\textwidth}
    \caption{Attention experts.}
    \label{tab:attention-experts}
    \centering
    \begin{tabular}{lc}
      \toprule
      Experts            & Accuracy ($\%$) \\
      \midrule
      Node (Transformer) & $71.7$          \\
      Node \& edge (GM)  & $86.0$          \\
      Edge               & $28.1$          \\
      \bottomrule
    \end{tabular}
  \end{minipage}
\end{table}

\paragraph{Attention experts}
Standard attention in Transformer uses only the node expert, whereas edge-augmented attention in GM combines the node and edge experts.
To complete the picture, we also test a variant that uses only the edge expert.
The edge-only variant underperforms the node-only variant, while both single-expert conditions underperform the GM double-expert condition, suggesting that node and edge experts provide complementary utility (Table~\ref{tab:attention-experts}).

\paragraph{Sharpener temperature}
We compare several parameterizations of the sharpener temperature: a fixed value shared across layers and edges, learned values per edge and layer, and projections from edge features at each layer.
Performance decreases under the alternative parameterizations, suggesting that feature-dependent sharpening is better in our setting (Table~\ref{tab:sharpener-temp}).

\begin{table}[!htbp]
  \caption{Sharpener temperature.}
  \label{tab:sharpener-temp}
  \centering
  \begin{tabular}{lcc}
    \toprule
    Sharpener temperature                      & Accuracy ($\%$) \\
    \midrule
    Fixed $=0$ (Transformer)                   & $71.7$          \\
    Fixed $=1$ - off                           & $77.5$          \\
    Fixed $=1.1$                               & $76.8$          \\
    Fixed $=1.5$                               & $72.5$          \\
    Learned values per edge                    & $80.8$          \\
    Learned projection from edge features (GM) & $86.0$          \\
    \bottomrule
  \end{tabular}
\end{table}

\paragraph{Factor temperatures}
Factor temperatures allow the model to modulate the relative contribution of node and edge experts.
Turning off factor temperatures reduces performance, suggesting that adaptive weighting between the two factors is beneficial (Table~\ref{tab:factor-temperatures}).

\begin{table}[!htbp]
  \centering
  \begin{minipage}{0.48\textwidth}
    \caption{Factor temperatures.}
    \label{tab:factor-temperatures}
    \centering
    \begin{tabular}{lc}
      \toprule
      Factor temperatures & Accuracy ($\%$) \\
      \midrule
      Off                 & $60.6$          \\
      On (GM)             & $86.0$          \\
      \bottomrule
    \end{tabular}
  \end{minipage}
  \hfill
  \begin{minipage}{0.48\textwidth}
    \caption{Entropy losses.}
    \label{tab:entropy-losses}
    \centering
    \begin{tabular}{lc}
      \toprule
      Entropy losses & Accuracy ($\%$) \\
      \midrule
      Off            & $84.5$          \\
      On (GM)        & $86.0$          \\
      \bottomrule
    \end{tabular}
  \end{minipage}
\end{table}

\paragraph{Entropy losses}
Preliminary experiments suggest that small entropy losses on the edge and node $n_2$ factors of both edge and node sublayers, each with weight starting at $10^{-3}$ and annealed linearly to $0$, reduce performance variability early in training.
We also observe a slight but non-significant improvement in final performance attributed to the entropy losses (Table~\ref{tab:entropy-losses}).

\subsection{Mechanistic analysis}
\label{sec:mechanistic-analysis}

\begin{figure}[!htbp]
  \centering
  \includegraphics[width=0.96\linewidth]{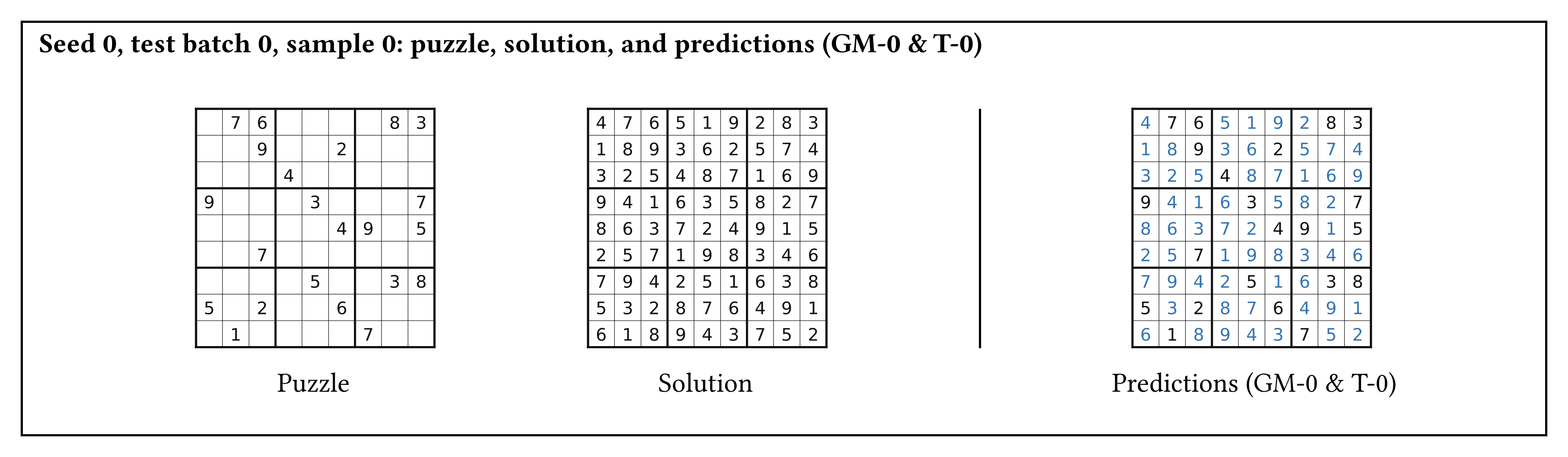}
  \caption{Seed $0$, test batch $0$, sample $0$: puzzle, solution, and predictions (GM-$0$ and T-$0$).}
  \label{fig:mechanistic-puzzle}
\end{figure}

The explicit edge representations and mechanisms lend GM to self-interpretability.
Various variables in each sublayer are weights over the target nodes, which in our experiments correspond to the $81$ Sudoku cells.
These variables include edge addresses, attention/referral edge/node factors, and attention/referral weights.
To visualize the representations and mechanisms, we record these variables during the testing of the final models and plot them as heatmaps.

We restrict this analysis to the GM and Transformer conditions from Section~\ref{sec:main-experiments}, using seed $0$, test batch $0$, and samples $0$--$15$.
We focus primarily on one pre-specified source cell, $r2c2$, located at the center of the top-left box, and extend the analysis to two additional cells, $r2c5$ and $r5c2$, where relevant.
Throughout this section, we use $0$-based indexing and refer to the two examined models as GM-$0$ and T-$0$.
We treat each consecutive pair of an edge sublayer and a node sublayer as a layer.
We use the term referral head to denote the new-edge-producing dimension during referral; $s$, $l$, and $e$ to denote sample, layer, and edge; $a$ and $r$ to denote attention and referral; and address-in to denote edge addresses after sharpening in either $a$ or $r$.
For example, referral head $s0l0r0$ produces an edge address that becomes two address-in variables after sharpening: $s0l0e1r$ in the edge sublayer and $s0l0e1a$ in the node sublayer.

Our analysis focuses mainly on sample $0$, which both GM-$0$ and T-$0$ solve correctly (Figure~\ref{fig:mechanistic-puzzle}).

\begin{figure}[!htbp]
  \centering
  \includegraphics[width=0.96\linewidth]{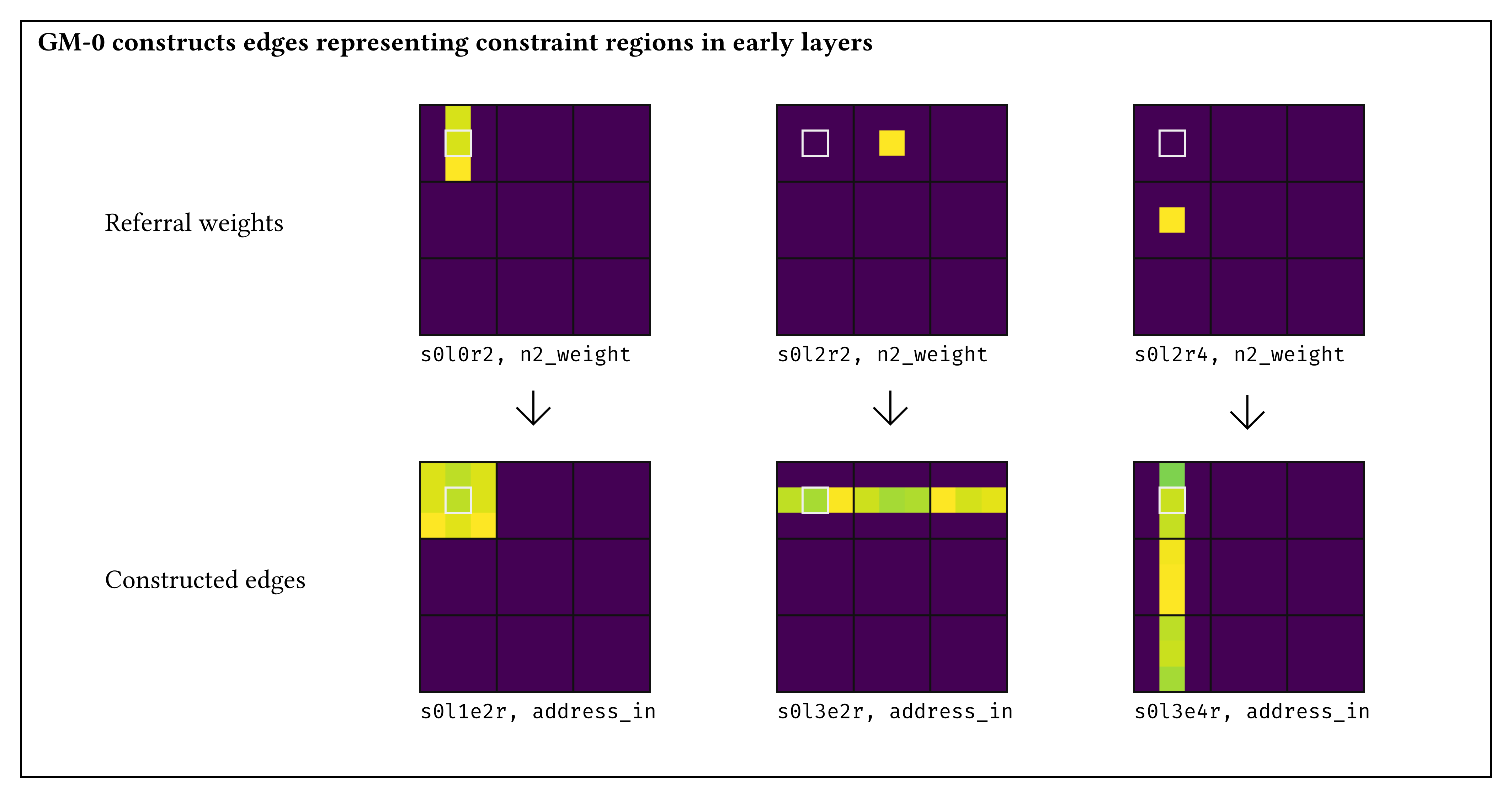}
  \caption{GM-$0$ constructs edges representing constraint regions in early layers.}
  \label{fig:mechanistic-gm-construct-constraints}
\end{figure}

In early layers, GM-$0$ builds Sudoku geometry from the input edges through a process resembling how convolutional neural networks compose elementary filters into more complex shapes~\citep{cnn-vis} (Figure~\ref{fig:mechanistic-gm-construct-constraints}).
For the examined node $r2c2$, referral head $2$ in layer $0$ uses the up, self, and down edges to construct an edge factor corresponding to a vertical $3\times 1$ rectangle containing $r1c2$, $r2c2$, and $r3c2$.
It then uses this factor to fetch the left and right edges of the up and down neighbors.
The fetched corner cells complete the box region representing the box constraint of $r2c2$.
Across all $16$ samples, the model consistently uses referral head $l0r2$ and the resulting $l1e2$ edges for the box constraint.

Similarly, the model obtains the $1\times 9$ rectangle corresponding to the row constraint region through the $l3e2$ edge, and the $9\times 1$ rectangle corresponding to the column constraint region through the $l3e4$ edge.
It does so by fetching edges from cells near the middle of the row or column, namely $r2c5$ or $r5c2$ (Figure~\ref{fig:mechanistic-gm-target-middle}).
These middle-of-row/column cells appear to serve as providers of the row and column regions for cells aligned with them (Figure~\ref{fig:mechanistic-gm-perform-algorithm}).
This is natural given their centrality: they are the only cells that can reach both ends of a row or column, and therefore provide the complete region by the third layer using edge traversal alone.

They can do so by recursively reaching for the farthest available cells in each direction, in coordination with other nodes.
At layer $0$, the $1$-distance neighbors are provided as input.
At layer $1$, the $2$-distance neighbors are fetched as the $1$-distance neighbors' $1$-distance neighbors.
At layer $2$, the $4$-distance neighbors are fetched as the $2$-distance neighbors' $2$-distance neighbors.
Although the model could in principle memorize each full row or column region, the observed pattern is consistent with a simpler and more general $1$-$2$-$4$ construction process shared across such nodes.
This exemplifies that, unlike standard GNNs, GMs have exponential rather than linear reach on the input graph given layer, and that explicit edge mechanisms may support stronger generalization under relational tasks.

Two additional phenomena merit discussion.
First, the partial edges formed during this process, namely the horizontal $l2e5$ edge and vertical $l2e1$ edge, also appear in source cells that are not middle-of-row/column cells (Figure~\ref{fig:mechanistic-gm-perform-algorithm}).
Thus, they appear to be byproducts of the $1$-$2$-$4$ construction applied broadly across nodes.
They may be fully useful only at middle-of-row/column cells, while remaining harmless or weakly beneficial elsewhere.

Second, GM-$0$ appears to locate the middle-of-row/column target cells through different mechanisms for rows and columns.
For the row region, it seems to rely on the $e2$ factor; for the column region, it relies on a joint contribution from the $n2$ edge and node factors (Figure~\ref{fig:mechanistic-gm-target-middle}).
This holds even when the target cell is the source cell itself, as in the case of $r2c5$ fetching its own edge for the row region.

\begin{figure}[!htbp]
  \centering
  \includegraphics[width=0.96\linewidth]{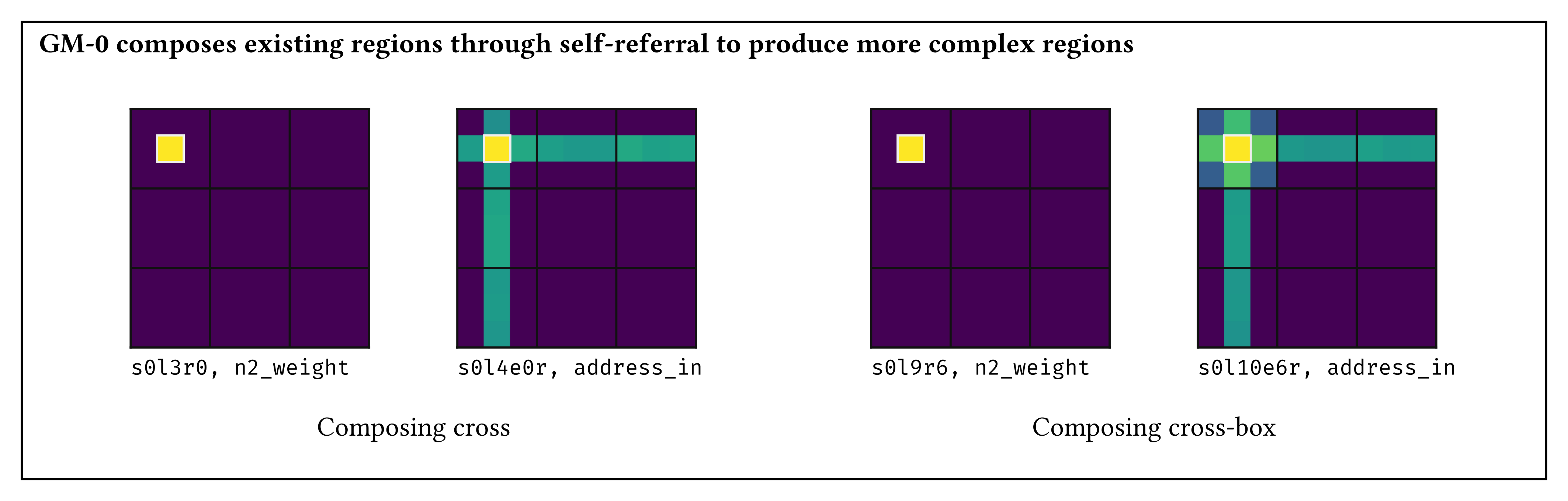}
  \caption{GM-$0$ composes existing regions through self-referral to produce more complex regions.}
  \label{fig:mechanistic-gm-compose-regions}
\end{figure}

The row and column regions are further composed into a cross region, represented by the $l4e0$ edge, which later combines with the box region to form a fused cross-box region (Figure~\ref{fig:mechanistic-gm-compose-regions}).
As permitted by the inclusion of self-edges, these regions are constructed and maintained through self-referral: the model retrieves its own previous edges using a referral edge factor concentrated on the source cell itself.

Unexpectedly, rather than preserving a self-edge across layers as the basis for self-referral, the model often adopts a more efficient strategy.
It composes existing non-self edges and applies a high edge-factor temperature to obtain a factor sharply concentrated on the source cell, thereby using one fewer edge slot while preserving the functional effect.
Across the $16$ samples, GM-$0$ generally does not maintain explicit self-edges beyond the second layer.

\begin{figure}[!htbp]
  \centering
  \includegraphics[width=0.96\linewidth]{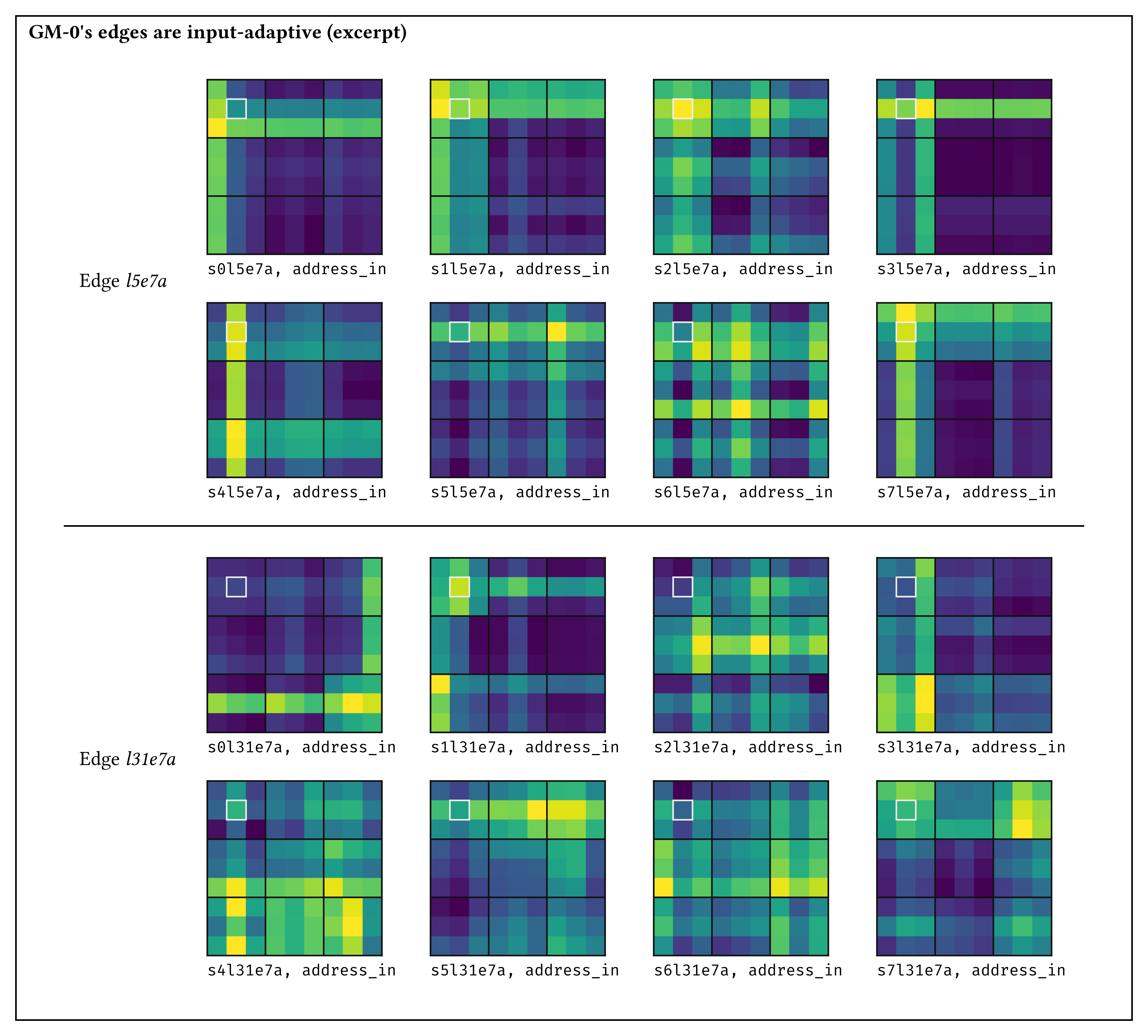}
  \caption{GM-$0$'s edges are input-adaptive (excerpt).}
  \label{fig:mechanistic-gm-adapt-edges-part}
\end{figure}

Apart from these fixed constructions, many of GM-$0$'s edges are input-adaptive,
suggesting that the model also represents abstract relations (Figure~\ref{fig:mechanistic-gm-adapt-edges-part}; additional examples are shown in Figures~\ref{fig:mechanistic-gm-adapt-edges-early} and~\ref{fig:mechanistic-gm-adapt-edges-late}).

\begin{figure}[!htbp]
  \centering
  \includegraphics[width=0.96\linewidth]{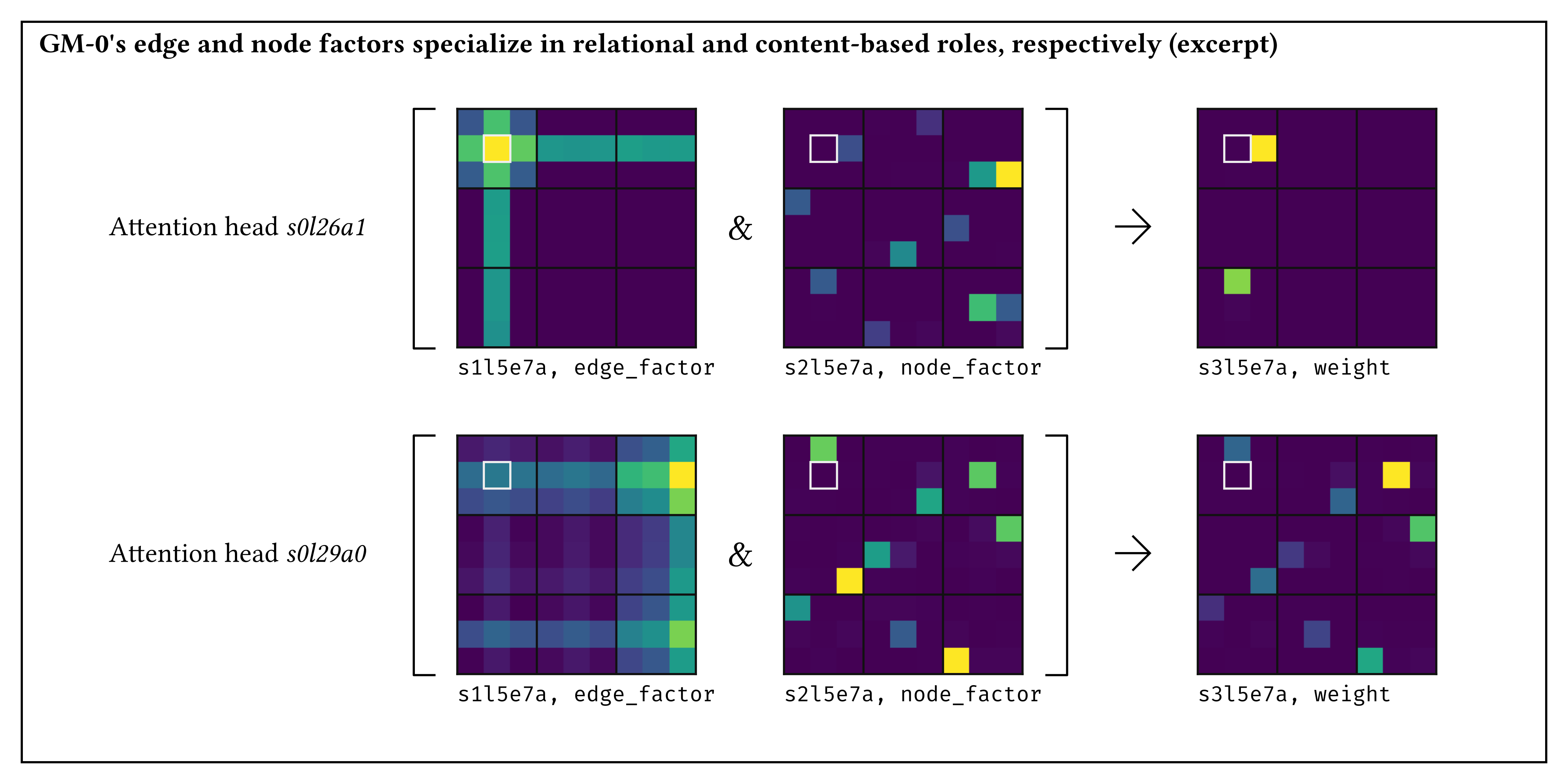}
  \caption{GM-$0$'s edge and node factors specialize in relational and content-based roles, respectively (excerpt).}
  \label{fig:mechanistic-gm-divide-labor-part}
\end{figure}

While GM-$0$'s edge factors seem to focus on relations, such as constraints, without encoding content preference, such as given or inferred digits, its node factors seem to match content without encoding relations.
This produces scattered patterns with no apparent geometric structure (Figure~\ref{fig:mechanistic-gm-divide-labor-part}; additional examples are shown in Figure~\ref{fig:mechanistic-gm-divide-labor}).
Often, the pattern of a node factor closely matches cells associated with a particular digit.

This suggests that, as intended, the model uses a division of labor between the two experts, which are then combined through a PoE to specify the attention or referral target.
In most cases, the resulting attention/referral weights appear to receive substantial contributions from both factors, although either expert can sometimes dominate.
Overall, across GM models, edge factors contribute more strongly to target specification:
their normalized entropies average $0.741$, compared with $0.891$ for node factors, during attention, and $0.402$, compared with $0.945$, during referral (Table~\ref{tab:factor-entropies}).

\begin{figure}[!htbp]
  \centering
  \includegraphics[width=0.96\linewidth]{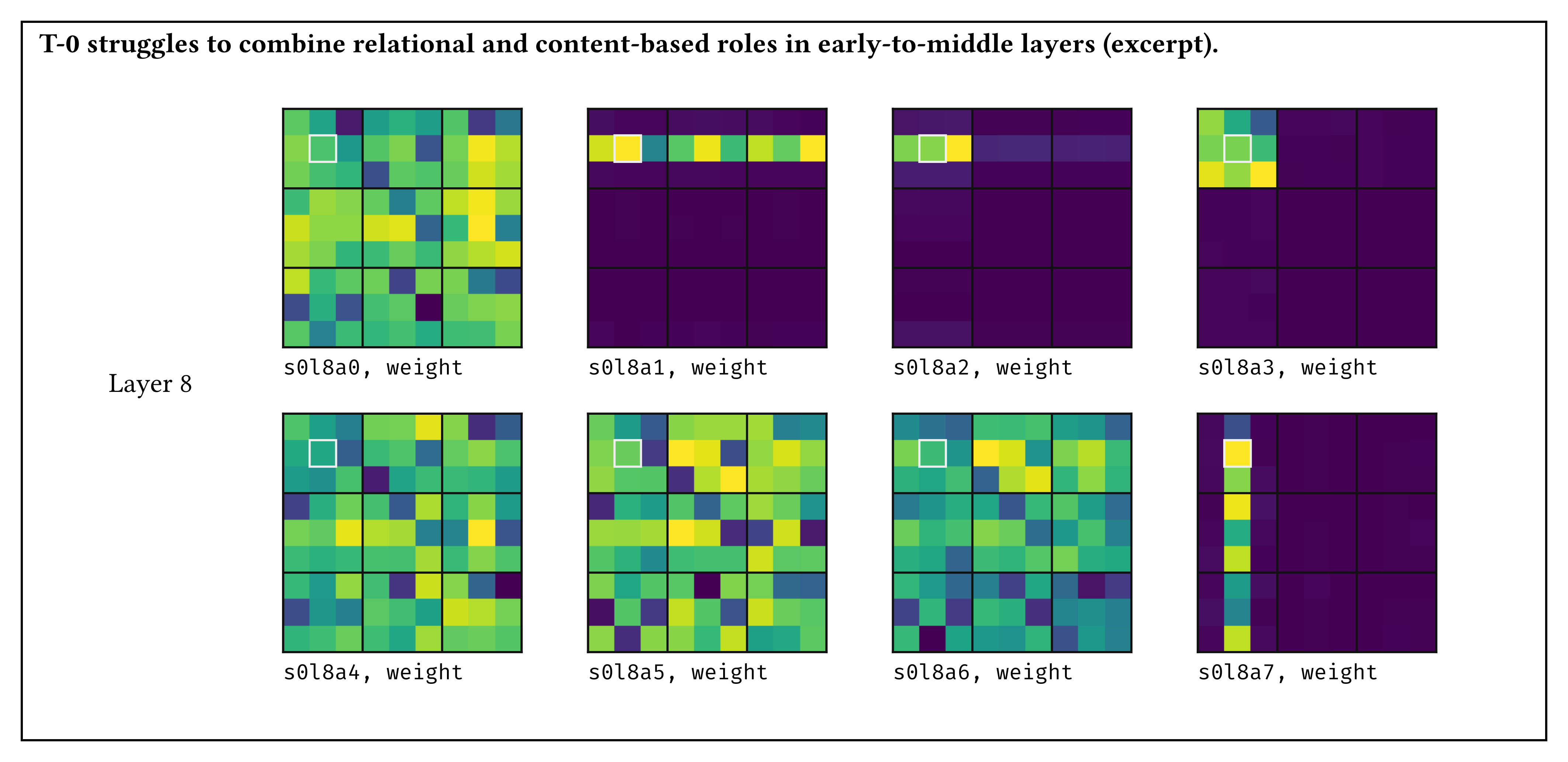}
  \caption{T-$0$ struggles to combine relational and content-based roles in early-to-middle layers (excerpt).}
  \label{fig:mechanistic-t-struggle-part}
\end{figure}

In contrast, T-$0$'s node factors, which are the sole contributors of attention weights, appear to struggle to fulfill both roles (Figure~\ref{fig:mechanistic-t-struggle-part}; additional examples are shown in Figure~\ref{fig:mechanistic-t-struggle}).
In early layers, although their node factors can attend based on both content and simple relational information, such as row, column, and box constraint regions,
they appear to struggle with relations beyond these simple groups, as well as with combining relational and content-based information.
In the final layers, T-$0$ appears to combine relational and content-based information in a primitive manner, although its attention remains less global and less interpretable than that of GM-$0$ (Figure~\ref{fig:mechanistic-gm-t-differ}).

Overall, attention in the Transformer models tends to be less sharp than in the GM models, with an average normalized entropy of $0.736$, compared with $0.652$ for GM (Table~\ref{tab:factor-entropies}).
Since our Sudoku instances can usually be solved using rudimentary ``Singles'' techniques~\citep{sudoku-singles}, these weaker attention patterns may still suffice for many examples, narrowing the performance gap with GM.
This highlights the need to test on tasks with a more complex relational structure.

\section{Related works}

Many Transformer and attention variants, such as Graphormer~\citep{graphormer}, incorporate notions of edges, relations, or graphs.
However, much of this work treats these as auxiliary to the node representation and mechanisms.
Graph Machine is instead situated among architectures that make edges a first-class component of the model through explicit and dynamic edge states.
Some representative examples include Edge Transformers~\citep{edge-transformer}, Edge-augmented Graph Transformers (EGT)~\citep{egt}, and relational attention~\citep{relational-attention}.
Among this category of architectures, GM differs along four dimensions.

First, GM preserves the node-centric Transformer pathway.
For example, our edge-augmented attention allows joint contribution from both node factors and edge factors.
This makes GM a more direct expressivity generalization of Transformers and enables it to apply naturally beyond graph-specific settings.

Second, GM uses a factorized edge state.
Many edge-state architectures maintain representations of shape $n \times n \times d_e$, assigning a dense vector to every ordered pair of nodes.
GM instead represents edges using edge features of shape $n \times k \times d_e$ and edge addresses of shape $n \times k \times n$.
This can be viewed as a generalization of the dense pairwise edge state: a dense $n \times n \times d_e$ representation corresponds to $n$ edge slots per node with one-hot addresses over destinations.
This separates what an edge contains from where it points, makes edge addresses directly interpretable as target mass, exposes edge degree as another controllable architectural inductive bias,
and clarifies ways to reduce the cost of edge referral through address compression or sparsification.

Third, GM differs in how edges shape information aggregation between nodes.
Rather than allowing edge representations to enter global content-based matching in an unconstrained manner, GM uses node and edge features to aggregate edge addresses, which then directly modulate attention.
This biases the edge pathway toward relational traversal, while leaving global content-based matching primarily to the node pathway.

Fourth, GM differs in how edge states are updated.
Alternative mechanisms do not explicitly support edge referral or regularize against rich contributions from the node pathway.
Our referral mixes edge addresses across node and edge-slot dimensions, such that new edge addresses are weighted combinations of previous edge addresses prior to sharpening.

Sudoku has also been used as a benchmark for relational and iterative reasoning.
Recurrent Relational Networks~\citep{rrn} inject a strong task prior by constructing a graph in which each cell is connected to cells in the same row, column, and box.
More recent recurrent reasoning architectures, including the Hierarchical Reasoning Model~\citep{hrm} and Tiny Recursive Model~\citep{trm}, evaluate on harder Sudoku variants using Transformer modules applied recurrently for many steps.
Our use of Sudoku is different: we do not provide a full hand-specified Sudoku constraint graph and only use a simple pass of the model.

\section{Conclusion and limitations}

We introduced Graph Machine, a neural architecture that represents graphs through both node features and dynamic edge representations.
By combining edge-augmented attention and edge-centric referral, the model can use soft graph structure during node updates while also revising that structure across layers.
In controlled Sudoku experiments, GM improves over standard-scale Transformer baselines and remains competitive with enlarged Transformer variants,
suggesting that explicit edge representations and mechanisms can improve relational reasoning.

The main limitation of the current implementation is computational cost.
Dense and uncompressed edge addresses yield cubic time complexity and quadratic memory footprint.
Future work should aim for more efficient referral mechanisms, including with sparse edge addresses and operations.
Because GM factorizes edge features and edge addresses, it provides a natural interface for such approximations.

Our evaluation is also limited to Sudoku.
Sudoku is useful as a controlled relational benchmark, but its latent graph structure is superficial, concrete, fixed, and regular.
This can favor Transformer baselines, since the latent structure can be relatively easily memorized via node embeddings.
GM may be more advantageous in settings where useful relations are latent, abstract, dynamic, and complex.

Additionally, Sudoku places unusually high demands on dense working memory: humans often require help from verbal rehearsal and pencil marks to solve Sudoku puzzles.
Thus, although edge address distributions are broad in our experiment, it remains worth examining whether this pattern persists in tasks with more typical working-memory demands.

More broadly, our results motivate continued exploration of edge mechanisms as architectural inductive biases.
We hypothesize that such mechanisms may help improve abstraction, reasoning, and generalization, capabilities that remain important challenges for current machine learning systems.
Graph Machine offers an initial step in this direction.

\clearpage

\bibliographystyle{unsrt}
\bibliography{graph-machine}

\clearpage

\appendix

\section{Simplified codes}

\subsection{Simplified code for edge-augmented attention}
\label{sec:simplified-code-attention}

\begin{figure}[h]
  \centering
  \begin{lstlisting}[language=Python, basicstyle=\scriptsize\ttfamily, tabsize=4, frame=lines, framesep=3mm, keepspaces=true, showspaces=false, showstringspaces=false, showtabs=false, columns=fullflexible]
def edge_augmented_attention(n1_queries, e1_keys, e1_addresses, n2_keys, n2_values):
	"""
	Args:
		n1_queries        (b, h, n,       d_ak): queries projected from n1's node features.
		edge_factor_temps (b, h, n            ): temps   projected from n1's node features.
		node_factor_temps (b, h, n            ): temps   projected from n1's node features.
		e1_keys           (b, h, n, K,    d_ak): keys    projected from e1's edge features.
		e1_addresses      (b,    n, K,    n   ):                        e1's edge addresses.
		n2_keys           (b, h,       n, d_ak): keys    projected from n2's node features.
		n2_values         (b, h,       n, d_av): values  projected from n2's node features.
	Rets:
		outs              (b, h, n,       d_av): attention outputs for  n1's node features.
	Einsum symbols:
		                  [b, h, n, k, m, a/d ]
	"""
	# Raw factors
	e1_scores        = einsum("bhnd,bhnkd->bhnk", n1_queries, e1_keys) / d_ak**0.5  # b, h, n, k
	e1_weights       = softmax(e1_scores, dim=-1)                                   # b, h, n, k
	edge_raw_factors = log(einsum("bhnk,bnkm->bhnm", e1_weights, e1_addresses))     # b, h, n, n
	node_raw_factors = einsum("bhnd,bhmd->bhnm", n1_queries, n2_keys) / d_ak**0.5   # b, h, n, n
	
	# Factors
	edge_factors = edge_raw_factors * edge_factor_temps.unsqueeze(-1)  # b, h, n, n
	node_factors = node_raw_factors * node_factor_temps.unsqueeze(-1)  # b, h, n, n
	
	# Weights
	logits  = edge_factors + node_factors  # b, h, n, n
	weights = softmax(logits, dim=-1)      # b, h, n, n

	# Outs
	outs = einsum("bhnm,bhmd->bhnd", weights, n2_values)  # b, h, n, d_av
	return outs
	\end{lstlisting}
  \caption{Simplified code for edge-augmented attention.}
  \label{fig:simplified-code-attention}
\end{figure}

\clearpage
\subsection{Simplified code for edge-centric referral}
\label{sec:simplified-code-referral}

\begin{figure}[h]
  \centering
  \begin{lstlisting}[language=Python, basicstyle=\scriptsize\ttfamily, tabsize=4, frame=lines, framesep=3mm, keepspaces=true, showspaces=false, showstringspaces=false, showtabs=false, columns=fullflexible]
def edge_centric_referral(n1_queries, ..., e2_addresses):
	"""
	Args:
		n1_queries           (b, k, n,          d_rk): queries projected from n1's node features.
		n2_edge_factor_temps (b, k, n               ): temps   projected from n1's node features.
		n2_node_factor_temps (b, k, n               ): temps   projected from n1's node features.
		e2_factor_temps      (b, k, n               ): temps   projected from n1's node features.
		e1_keys              (b, k, n, k,       d_rk): keys    projected from e1's edge features.
		e1_values            (b, k, n, k,       d_e ): values  projected from e1's edge features.
		e1_addresses         (b,    n, k,       n   ):                        e1's edge addresses.
		n2_keys              (b, k,       n,    d_rk): keys    projected from n2's node features.
		n2_values            (b, k,       n,    d_e ): values  projected from n2's node features.
		e2_keys              (b, k,       n, k, d_rk): keys    projected from e2's edge features.
		e2_values            (b, k,       n, k, d_e ): values  projected from e2's edge features.
		e2_addresses         (b,          n, k, n   ):                        e2's edge addresses.
	Rets:
		feature_outs         (b, k, n,          d_e ): referral outputs  for  e1's edge features.
		address_outs         (b, k, n,          n   ): referral outputs  for  e1's edge addresses.
	Einsum symbols:
		                     [b, h, n, k, m, l, a/d ]
	"""
	# Raw factors
	*_, d_rk  = n1_queries.shape
	e1_scores           = einsum("bhnd,bhnkd->bhnk",  n1_queries, e1_keys) / d_rk**0.5  # b, k, n, k
	e1_weights          = softmax(e1_scores, dim=-1)                                    # b, k, n, k
	n2_edge_raw_factors = log(einsum("bhnk,bnka->bhna", e1_weights, e1_addresses))      # b, k, n, n
	n2_node_raw_factors = einsum("bhnd,bhmd->bhnm",   n1_queries, n2_keys) / d_rk**0.5  # b, k, n, n
	e2_raw_factors      = einsum("bhnd,bhmld->bhnml", n1_queries, e2_keys) / d_rk**0.5  # b, k, n, n, k

	# Factors
	n2_edge_factors = n2_edge_raw_factors * n2_edge_factor_temps.unsqueeze(-1)           # b, k, n, n
	n2_node_factors = n2_node_raw_factors * n2_node_factor_temps.unsqueeze(-1)           # b, k, n, n
	e2_factors      = e2_raw_factors      * e2_factor_temps.unsqueeze(-1).unsqueeze(-1)  # b, k, n, n, k
	n2_factors      = n2_edge_factors + n2_node_factors                                  # b, k, n, n

	# Weights
	e2_logits  = n2_factors.unsqueeze(-1) + e2_factors  # b, k, n, n, k
	e2_weights = softmax(dim=(-1, -2))                  # b, k, n, n, k
	n2_weights = e2_weights.sum(dim=-1)                 # b, k, n, n

	# Outs
	e1_feature_outs = einsum("bhnk,bhnkd->bhnd",  e1_weights, e1_values)     # b, k, n, d_e
	n2_feature_outs = einsum("bhnm,bhmd->bhnd",   n2_weights, n2_values)     # b, k, n, d_e
	e2_feature_outs = einsum("bhnml,bhmld->bhnd", e2_weights, e2_values)     # b, k, n, d_e
	feature_outs    = e1_feature_outs + n2_feature_outs + e2_feature_outs    # b, k, n, d_e
	address_outs    = einsum("bhnml,bmla->bhna",  e2_weights, e2_addresses)  # b, k, n, n
	return feature_outs, address_outs
	\end{lstlisting}
  \caption{Simplified code for edge-centric referral.}
  \label{fig:simplified-code-referral}
\end{figure}

\clearpage

\section{Default configurations}
\label{sec:default-config}

For our default configurations, we follow common Transformer design choices, using RMSNorm~\citep{rms}, no bias, and SwiGLU~\citep{swiglu} activation function.
Both the feed-forward networks in the edge and node sublayers have an expansion factor of $4$.
We use as our temperature function a horizontally shifted softplus that maps $0$ to $1$.

Training uses PyTorch automatic mixed precision via \texttt{torch.autocast}.
Adam optimizer~\citep{adam} is used with $(\beta_1,\beta_2)$ of $(0.9,0.95)$, no weight decay, a $1\%$ warmup, a peak learning rate of $10^{-3}$, and a cosine schedule down to $10\%$.
Gradient clipping with max grad norm of $1.0$ is applied.

Train-eval-test split on the dataset is redrawn for each run with proportions $0.9$-$0.05$-$0.05$.
$100k$ steps are performed with batch size of $64$ each, corresponding to $2.37$ epochs over the train set; each epoch reshuffles the dataset.
Testing is done with the entirety of the test split at the end of each run.

\section{Additional experiments and statistics}

\paragraph{Transformer - project input edges}
\label{sec:transformer-proj}
We test a Transformer condition where we project the initial edge embeddings and edge addresses to the node feature space in the input layer.
We do so by
\begin{enumerate}
  \item perform weighted sum of the learned node embeddings using the initial edge addresses as weights for each edge,
  \item concatenate with the edge embeddings for each edge,
  \item pass through an MLP for each edge,
  \item and add to the node embeddings.
\end{enumerate}
Results (Table~\ref{tab:transformer-proj}) suggest this procedure is not advantageous in our case.

\begin{table}[!htbp]
  \caption{Transformer - project input edges.}
  \label{tab:transformer-proj}
  \centering
  \begin{tabular}{llc}
    \toprule
    Base configuration & Project input edges & Accuracy ($\%$) \\
    \midrule
    \multirow{2}{*}{Transformer}
                       & On                  & $60.1$          \\
                       & Off (Transformer)   & $71.7$          \\
    \bottomrule
  \end{tabular}
\end{table}

\paragraph{Transformer-sin-PE-2x - factor temperatures}
\label{sec:transformer-sin-pe-2x-factor-temperatures}
On top of the Transformer-sin-PE-2x condition, we test the effect of removing factor temperatures.
Results (Table~\ref{tab:transformer-sin-pe-2x-factor-temperatures}) suggest that this seems to slightly improve performance.

\begin{table}[!htbp]
  \caption{Transformer-sin-PE-2x - factor temperatures.}
  \label{tab:transformer-sin-pe-2x-factor-temperatures}
  \centering
  \begin{tabular}{llc}
    \toprule
    Base configuration & Factor temperatures        & Accuracy ($\%$) \\
    \midrule
    \multirow{2}{*}{Transformer-sin-PE-2x}
                       & On (Transformer-sin-PE-2x) & $86.3$          \\
                       & Off                        & $83.7$          \\
    \bottomrule
  \end{tabular}
\end{table}

\paragraph{Transformer-sin-PE-2x - entropy losses}
\label{sec:transformer-sin-pe-2x-entropy-losses}
On top of the Transformer-sin-PE-2x condition, we test the effect of removing entropy losses.
Results (Table~\ref{tab:transformer-sin-pe-2x-factor-temperatures}) suggest that this does not seem to materially alter performance.

\begin{table}[!htbp]
  \caption{Transformer-sin-PE-2x - entropy losses.}
  \label{tab:transformer-sin-pe-2x-entropy-losses}
  \centering
  \begin{tabular}{llc}
    \toprule
    Base configuration & Entropy losses             & Accuracy ($\%$) \\
    \midrule
    \multirow{2}{*}{Transformer-sin-PE-2x}
                       & On (Transformer-sin-PE-2x) & $86.3$          \\
                       & Off                        & $87.6$          \\
    \bottomrule
  \end{tabular}
\end{table}

\paragraph{Transformer - positional encodings}
\label{sec:transformer-pe}
We test a range of $2$D positional encoding options, including RoPE of various bases, sinusoidal PE of various bases,
as well as a learned row-column factorized PE, where a node inherits the sum of the learned embeddings of its row and column.
Our experiments suggest that sinusoidal PE with a base of $10000$ performs the best among our conditions (Table~\ref{tab:transformer-sin-pe-2x-factor-temperatures}).
Among RoPE conditions, base $10$ is best, which we label as Transformer-RoPE to offer comparison with GM-RoPE in Section~\ref{sec:ablation-studies}.

\begin{table}[!htbp]
  \caption{Transformer - positional encodings.}
  \label{tab:transformer-pe}
  \centering
  \begin{tabular}{llc}
    \toprule
    Base configuration & Positional encodings                             & Accuracy ($\%$) \\
    \midrule
    \multirow{10}{*}{Transformer}
                       & Off (Transformer)                                & $71.7$          \\
                       & RoPE, base $10$ [Transformer-RoPE]               & $75.9$          \\
                       & RoPE, base $100$                                 & $67.7$          \\
                       & RoPE, base $1000$                                & $74.1$          \\
                       & RoPE, base $10000$                               & $71.2$          \\
                       & Sinusoidal PE, base $10$                         & $77.1$          \\
                       & Sinusoidal PE, base $100$                        & $80.3$          \\
                       & Sinusoidal PE, base $1000$                       & $80.2$          \\
                       & Sinusoidal PE, base $10000$ (Transformer-sin-PE) & $81.5$          \\
                       & Learned row-column factorized PE                 & $74.6$          \\
    \bottomrule
  \end{tabular}
\end{table}

\paragraph{GM - edge address space}
\label{sec:gm-edge-address-space}
We test a variant of GM where we let edge addresses represent logits directly, by removing the conversion to and from weight space when logits are needed.
During input, to convert the input edge-address weights to logits, we simply multiply the weights by a chosen factor of $5$.
As a result, the uniform weights of the empty edges convert to uniform logits and thus distribution,
and edge addresses with a single-cell support convert to a distribution where the mass on the intended cell is $e^5\approx 148.41$ larger than that of any other cell.
Results (Table~\ref{tab:gm-edge-address-space}) indicate that this reduces the performance to around Transformer level.

\begin{table}[!htbp]
  \caption{Edge address space.}
  \label{tab:gm-edge-address-space}
  \centering
  \begin{tabular}{llc}
    \toprule
    Base configuration & Edge address space & Accuracy ($\%$) \\
    \midrule
    \multirow{2}{*}{GM}
                       & Weight (GM)        & $86.0$          \\
                       & Logit              & $72.3$          \\
    \bottomrule
  \end{tabular}
\end{table}

\paragraph{GM - address sparsity}
\label{sec:gm-address-sparsity}
We conduct preliminary experiments on address sparsity by varying sparsity $s$,
simulated using our dense algorithm and representation by masking non-top-$s$ entries to $0$ along the last dimension of edge addresses.
We optionally use a Gumbel-top-$s$ temperature $\tau$ for our selection by adding Gumbel noise to our address logits before selecting the top $s$ of the original logits and renormalizing.
In our implementation, we do not further sparsify the edge factors or mask the attention/referral weights to maintain the top-$s$ sparsity.
We find that sparsification yields promising results, where a sparsity of $16$ yeilds similar performance to Transformer; however, using Gumbel-top-$s$ without annealing seems to yield no benefit (Table~\ref{tab:gm-address-sparsity}).
What remains future work includes enforcing sparsity on attention/referral weights for potentially significant efficiency benefits, and where we anneal $s$ and/or $\tau$ for potentially improved optimization.

\begin{table}[!htbp]
  \caption{GM - address sparsity.}
  \label{tab:gm-address-sparsity}
  \centering
  \begin{tabular}{lllc}
    \toprule
    Base configuration & $s$               & $\tau$ & Accuracy ($\%$) \\
    \midrule
    \multirow{8}{*}{GM}
                       & $0$ (Transformer) & $0.0$  & $71.7$          \\
                       & $1$               & $0.0$  & $36.2$          \\
                       & $4$               & $0.0$  & $50.9$          \\
                       & $16$              & $0.0$  & $70.3$          \\
                       & $81$ (GM)         & $0.0$  & $86.0$          \\
    \cmidrule{2-4}
                       & $4$               & $0.01$ & $27.5$          \\
                       & $4$               & $0.1$  & $30.4$          \\
                       & $4$               & $1.0$  & $53.6$          \\
    \bottomrule
  \end{tabular}
\end{table}

\paragraph{GM and Transformer - some statistics}
\label{sec:gm-t-stats}
We offer some statistics on the GM and Transformer conditions.
Factor entropies are logged after applying the factor temperatures.
In-addresses are the input addresses after sharpening, while out-addresses are the output addresses after the referral process.

\begin{table}[!htbp]
  \caption{Mean sharpener temperatures.}
  \label{tab:sharpener-temps}
  \centering
  \begin{tabular}{lcc}
    \toprule
    Condition & Edge sublayer & Node sublayer \\
    \midrule
    GM        & $1.050$       & $1.098$       \\
    \bottomrule
  \end{tabular}
\end{table}

\begin{table}[!htbp]
  \caption{Mean factor temperatures.}
  \label{tab:factor-temps}
  \centering
  \begin{tabular}{lccccc}
    \toprule
    \multirow{2}{*}[-0.5ex]{Condition} & \multicolumn{3}{c}{Edge sublayer} & \multicolumn{2}{c}{Node sublayer}                                             \\
    \cmidrule(lr){2-4} \cmidrule(lr){5-6}
                                       & N2 edge                           & N2 node                           & E2             & Edge           & Node    \\
    \midrule
    Transformer                        & \textbackslash                    & \textbackslash                    & \textbackslash & \textbackslash & $0.871$ \\
    GM                                 & $15.374$                          & $0.287$                           & $3.386$        & $1.096$        & $0.861$ \\
    \bottomrule
  \end{tabular}
\end{table}

\begin{table}[!htbp]
  \caption{Factor mean normalized entropies.}
  \label{tab:factor-entropies}
  \centering
  \begin{tabular}{lcccccc@{\;}c@{\;}c}
    \toprule
    \multirow{2}{*}[-0.5ex]{Condition} & \multicolumn{4}{c}{Edge sublayer} & \multicolumn{4}{c}{Node sublayer}                                                                                            \\
    \cmidrule(lr){2-5} \cmidrule(lr){6-9}
                                       & N2 edge                           & N2 node                           & E2             & Weight (N2)    & Edge           & Node    &                   & Weight  \\
    \midrule
    Transformer                        & \textbackslash                    & \textbackslash                    & \textbackslash & \textbackslash & \textbackslash & $0.736$ & \hspace{0.2em}$=$ & $0.736$ \\
    GM                                 & $0.402$                           & $0.945$                           & $0.731$        & $0.368$        & $0.741$        & $0.891$ &                   & $0.652$ \\
    \bottomrule
  \end{tabular}
\end{table}

\begin{table}[!htbp]
  \caption{Address mean normalized entropies.}
  \label{tab:address-entropies}
  \centering
  \begin{tabular}{lccc}
    \toprule
    \multirow{2}{*}[-0.5ex]{Condition} & \multicolumn{2}{c}{Edge sublayer} & \multicolumn{1}{c}{Node sublayer}              \\
    \cmidrule(lr){2-3} \cmidrule(lr){4-4}
                                       & In-address                        & Out-address                       & In-address \\
    \midrule
    GM                                 & $0.692$                           & $0.712$                           & $0.715$    \\
    \bottomrule
  \end{tabular}
\end{table}

\section{Additional mechanistic analysis figures}

This section collects some additional mechanistic analysis figures.

\begin{figure}[!htbp]
  \centering
  \includegraphics[width=0.96\linewidth]{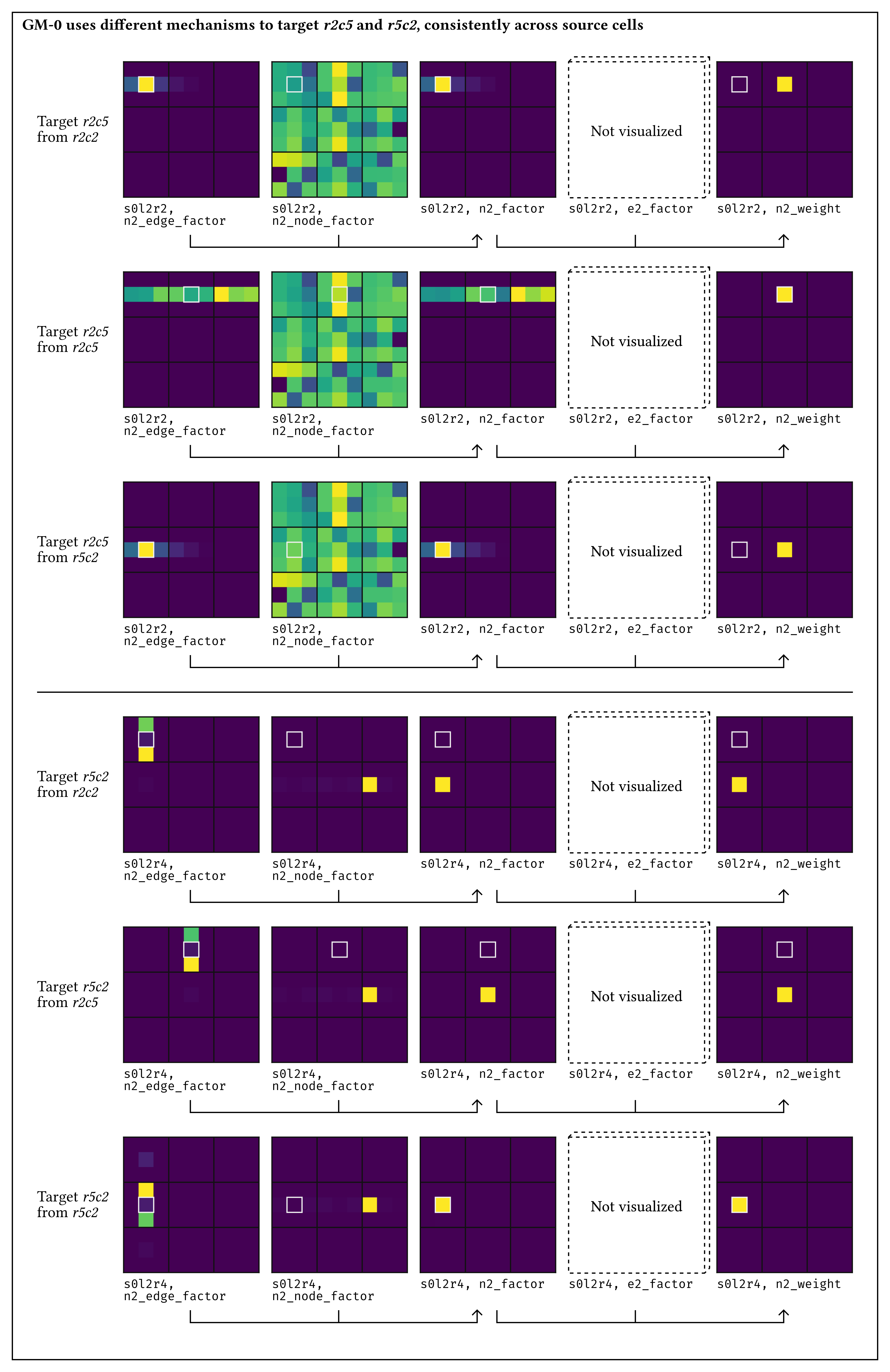}
  \caption{GM-$0$ uses different mechanisms to target $r2c5$ and $r5c2$, consistently across source cells.}
  \label{fig:mechanistic-gm-target-middle}
\end{figure}

\begin{figure}[!htbp]
  \centering
  \includegraphics[width=0.96\linewidth]{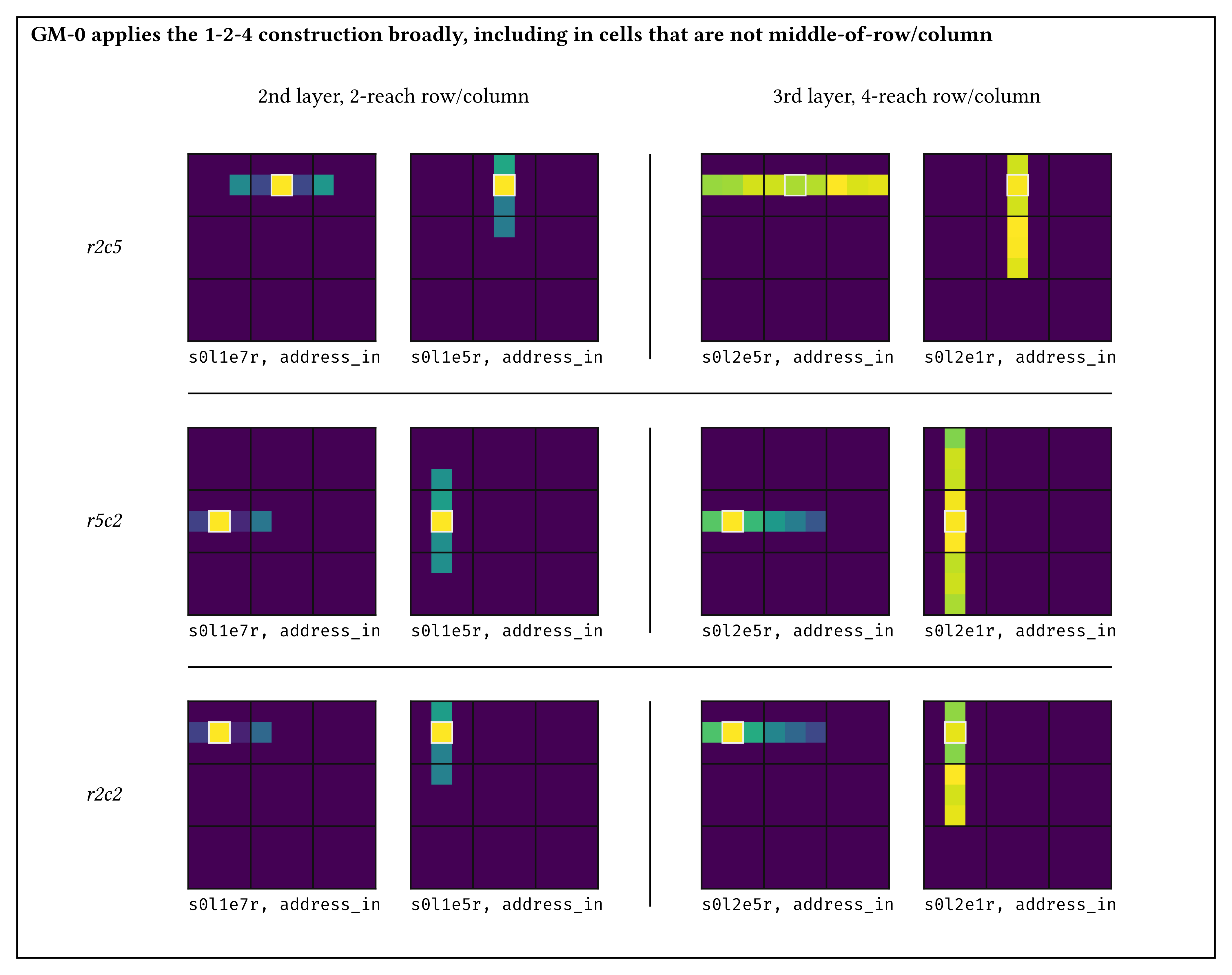}
  \caption{GM-$0$ applies the $1$-$2$-$4$ construction broadly, including in cells that are not middle-of-row/column.}
  \label{fig:mechanistic-gm-perform-algorithm}
\end{figure}

\begin{figure}[!htbp]
  \centering
  \includegraphics[width=0.96\linewidth]{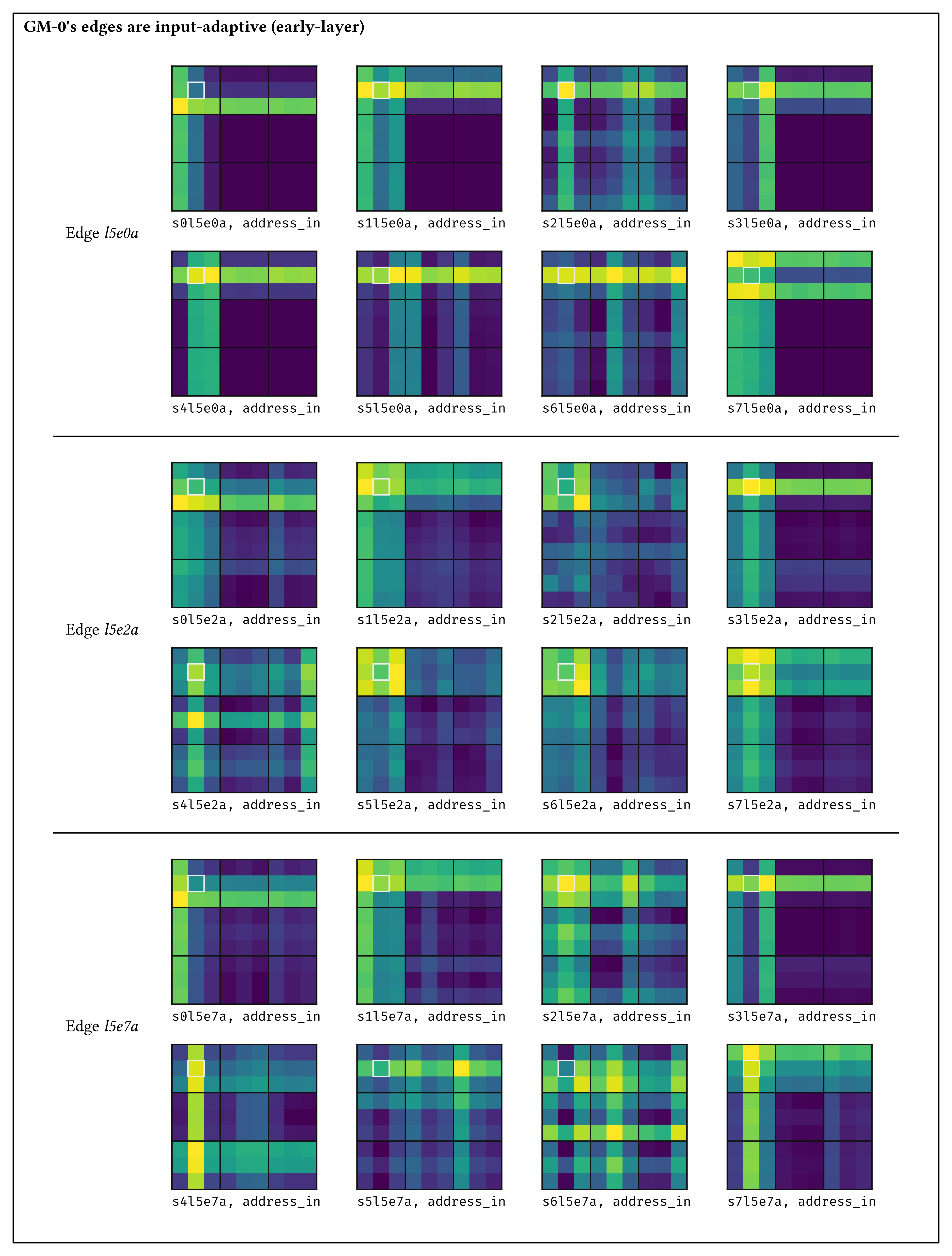}
  \caption{GM-$0$'s edges are input-adaptive (early-layer).}
  \label{fig:mechanistic-gm-adapt-edges-early}
\end{figure}

\begin{figure}[!htbp]
  \centering
  \includegraphics[width=0.96\linewidth]{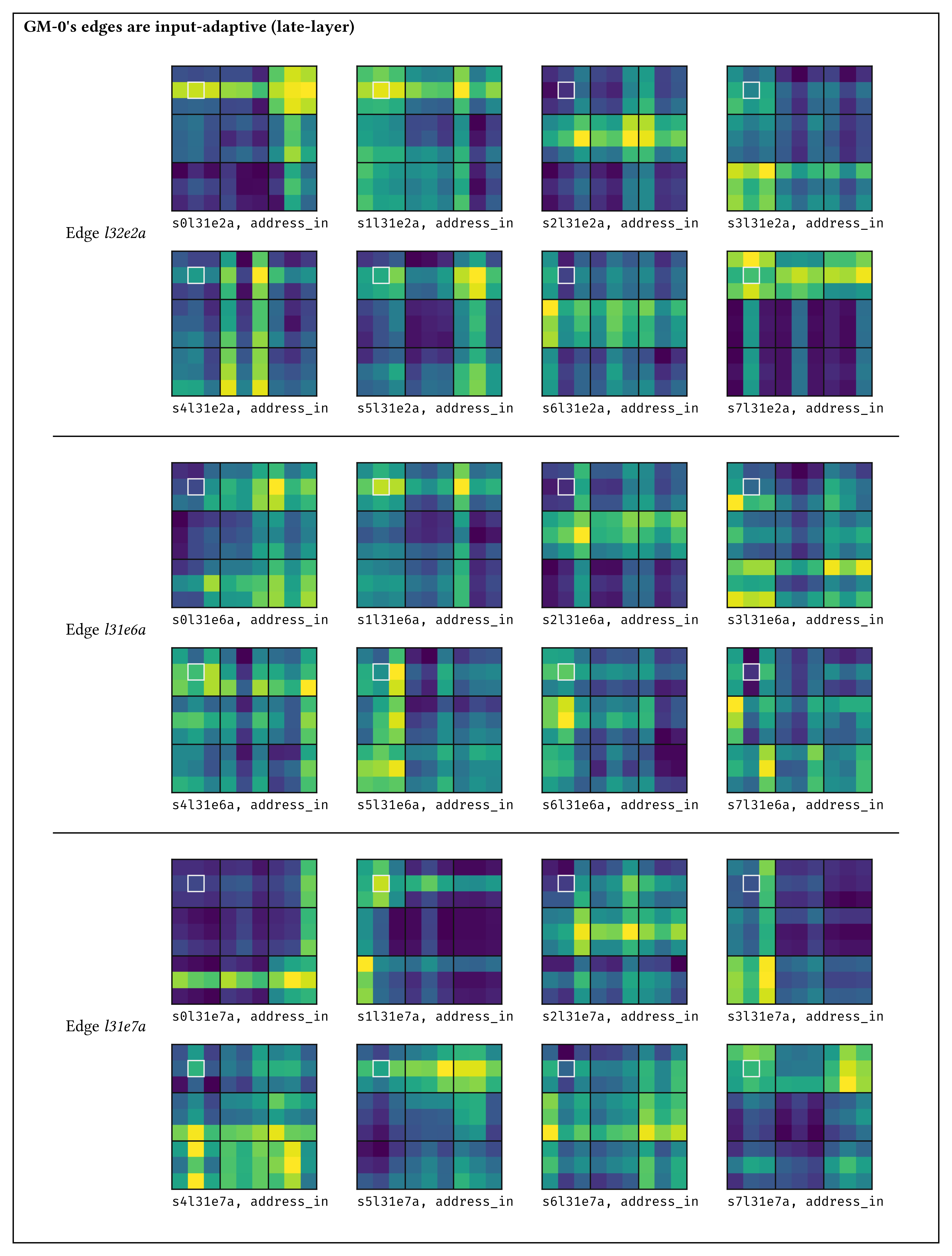}
  \caption{GM-$0$'s edges are input-adaptive (late-layer).}
  \label{fig:mechanistic-gm-adapt-edges-late}
\end{figure}

\begin{figure}[!htbp]
  \centering
  \includegraphics[width=0.96\linewidth]{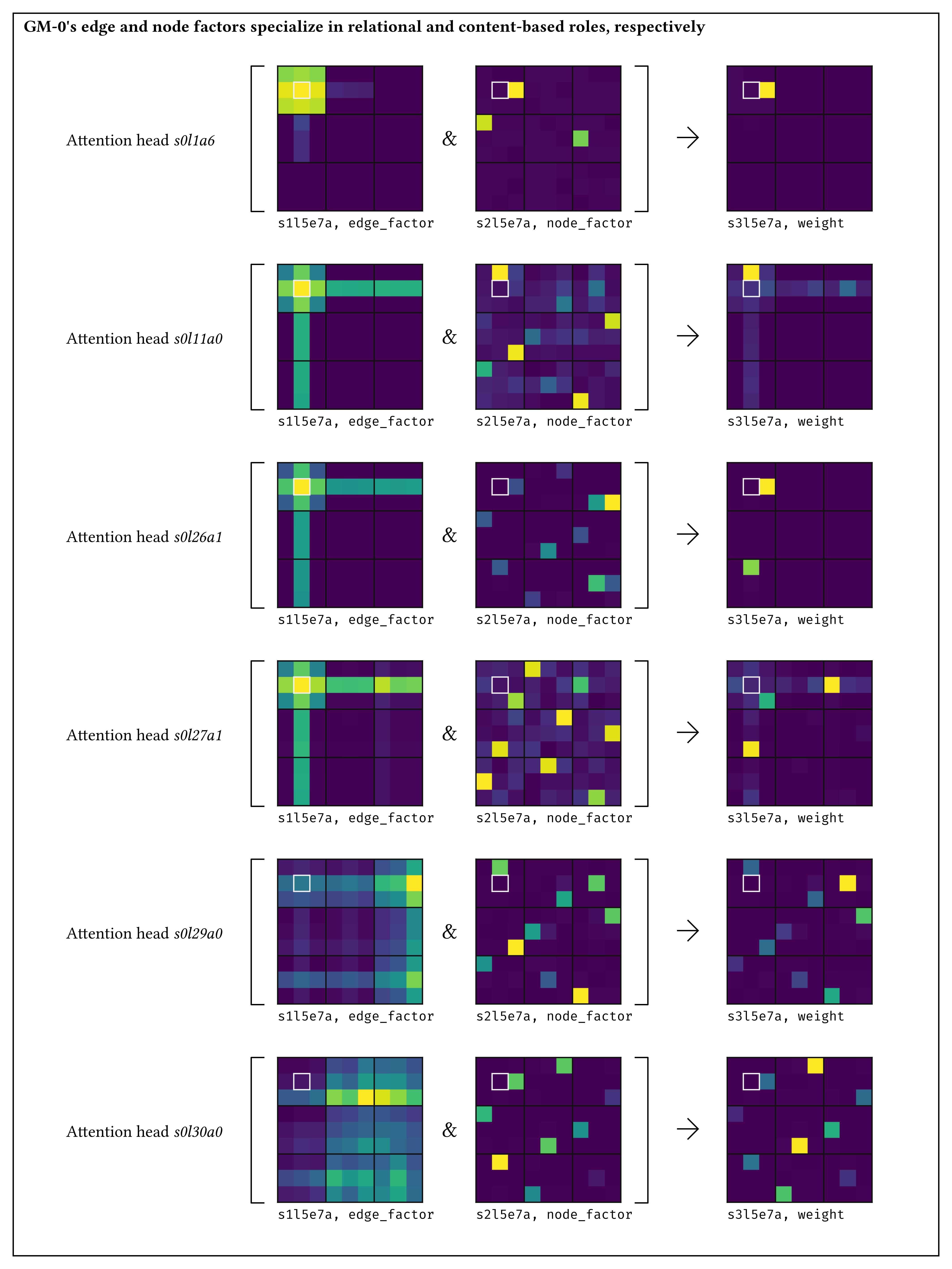}
  \caption{GM-$0$'s edge and node factors specialize in relational and content-based roles, respectively.}
  \label{fig:mechanistic-gm-divide-labor}
\end{figure}

\begin{figure}[!htbp]
  \centering
  \includegraphics[width=0.96\linewidth]{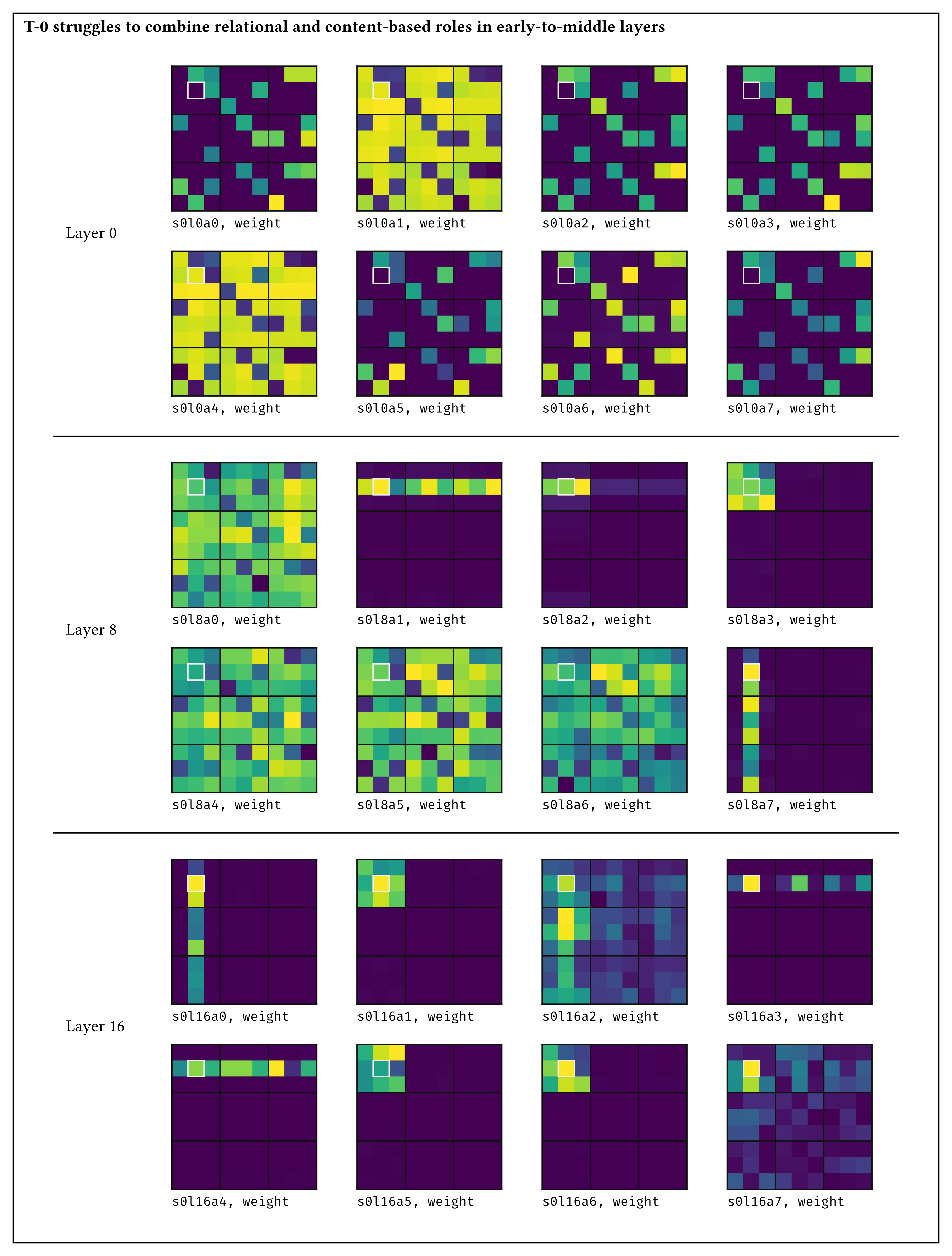}
  \caption{T-$0$ struggles to combine relational and content-based roles in early-to-middle layers.}
  \label{fig:mechanistic-t-struggle}
\end{figure}

\begin{figure}[!htbp]
  \centering
  \includegraphics[width=0.96\linewidth]{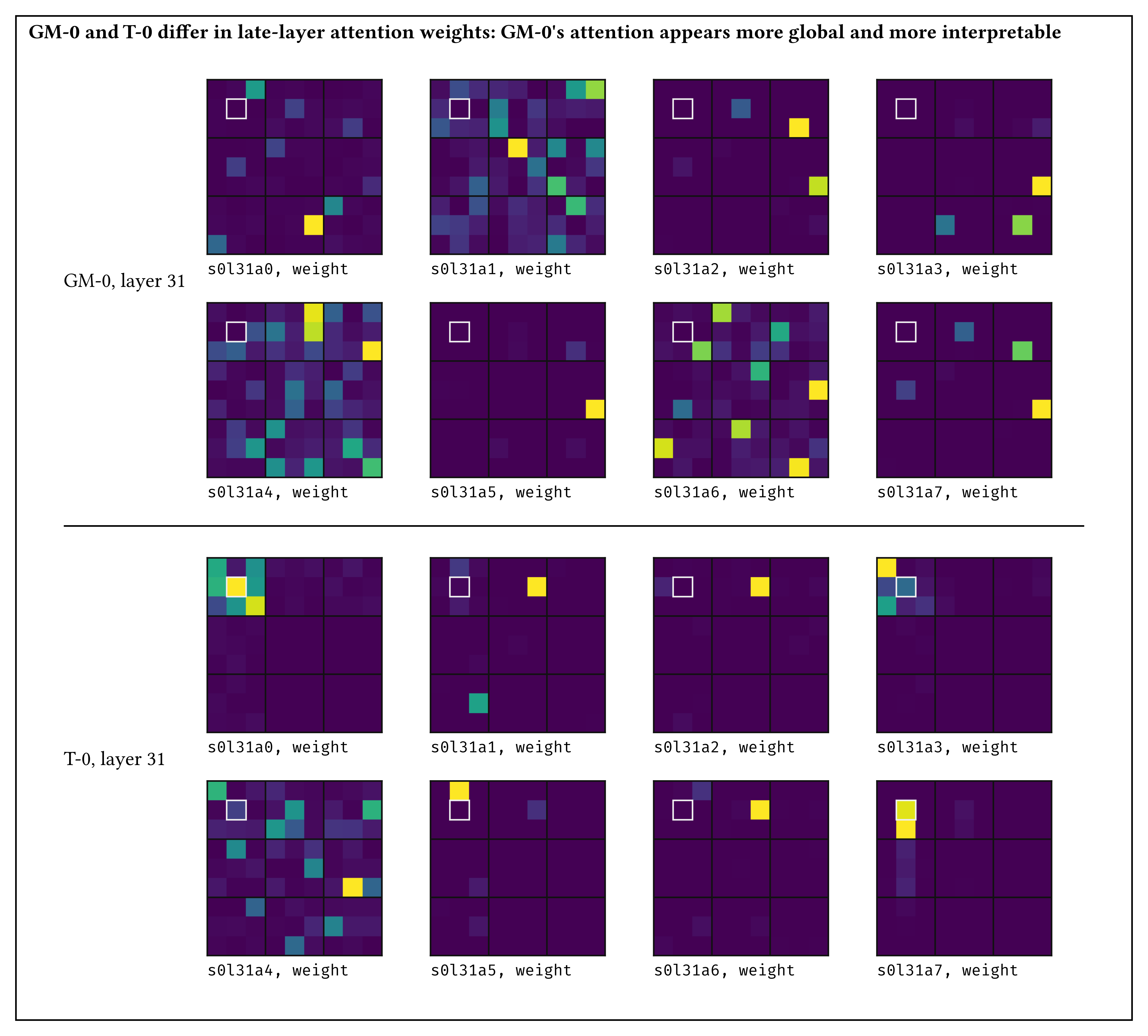}
  \caption{GM-$0$ and T-$0$ differ in late-layer attention weights: GM-$0$'s attention appears more global and more interpretable.}
  \label{fig:mechanistic-gm-t-differ}
\end{figure}

\end{document}